\documentclass[preprint,11pt]{elsarticle}

\usepackage{amsmath,graphicx}
\usepackage[dvipsnames]{xcolor}
\usepackage{tikz}
\usetikzlibrary{shapes,calc,matrix}
\usetikzlibrary{matrix,chains,positioning,decorations.pathreplacing,arrows}
\usetikzlibrary{arrows.meta, chains, positioning, quotes}
\usetikzlibrary{positioning,calc}
\usepackage[british]{babel}
\usepackage[utf8]{inputenc}
\usepackage[normalem]{ulem}
\usepackage[T1]{fontenc}
\usepackage{mathtools}
\usepackage[thinc]{esdiff}

\usepackage{url}
\usepackage{amsfonts}
\usepackage{subfig}
\usepackage{nicefrac}
\usepackage{amsthm}
\usepackage{amssymb}

\usepackage{enumitem}
\usepackage{bm}
\usepackage{stfloats}
\usepackage[normalem]{ulem}
\usepackage{systeme}

\usepackage{algorithm, float}
\usepackage{algorithmic} 
\usepackage[squaren, Gray]{SIunits}
\usepackage[acronym,toc,nonumberlist, nopostdot, style=super, nogroupskip]{glossaries}
\usepackage{hyperref}
\usepackage{cuted}
\usepackage[section]{placeins}
\usepackage{tikz}
\usepackage{graphicx}
\usepackage{pgfplots}
\usetikzlibrary{positioning}
\tikzset{figNode/.style={ 
    path picture={
      \node at (path picture bounding box.center) {#1};}}
}
\tikzset{mySimpleArrow/.style n args={2}{
    >={latex[#1]},
    every path/.style={draw=#2}
  },
  mySimpleArrow/.default={gray}{gray}
}
\tikzset{myBlock/.style n args={2}{
    every node/.style={rectangle, text=black,
      minimum width=#1, minimum height=#2,}
  },
  myBlock/.default={1cm}{1cm}
}
\DeclareMathOperator*{\argreduce}{arg\,reduce}
\makeatletter
\renewcommand\p@subfigure{\thefigure\,}
\makeatother

\newtheorem{remark}{Remark} 

\pgfplotsset{compat=1.17}
\biboptions{sort&compress}

\newcommand{\revise}[1]{{{\color{black} #1}}}

\begin{document}

\begin{frontmatter}



\title{A consistent and flexible framework for \\ deep matrix factorizations}


\author[inst1]{Pierre De Handschutter\corref{cor1}} \ead{pierre.dehandschutter@umons.ac.be}
\author[inst1]{Nicolas Gillis} \ead{nicolas.gillis@umons.ac.be}
\affiliation[inst1]{organization={Department of Mathematics and Operational  Research \\ University of Mons},%Department and Organization
            country={Belgium}}
\cortext[cor1]{Corresponding author}

\begin{abstract}
Deep matrix factorizations (deep MFs) are recent unsupervised data mining techniques inspired by constrained low-rank approximations. They aim to extract complex hierarchies of features within high-dimensional datasets. 
Most of the loss functions proposed in the literature to evaluate the quality of deep MF models and the underlying optimization frameworks are not consistent because different losses are used at different layers. In this paper, we introduce two meaningful loss functions for deep MF and present a generic framework to solve the corresponding optimization problems. We illustrate the effectiveness of this approach through the integration of various constraints and regularizations, such as sparsity, nonnegativity and minimum-volume. The models are successfully applied on both synthetic and real data, namely for hyperspectral unmixing and extraction of facial features. 
\end{abstract}


%\item Two new loss functions for deep matrix factorizations are introduced
%
%\item Our loss functions are weighted sums of layer-based errors 
%
%\item A general convergent optimization framework is presented to tackle our formulations
%
%\item Several priors are incorporated, e.g., sparsity, nonnegativity and \\ \mbox{minimum-volume} 
%
%\item Experiments on synthetic and real data validate the effectiveness of our framework


\end{frontmatter}


\section{Introduction}
\label{sec:intro}

In the era of data science, the extraction of meaningful features in datasets is a crucial challenge. To do so, a fundamental class of unsupervised linear dimensionality reduction methods is low-rank matrix factorizations~(LRMFs). 
Given a matrix $X \in \mathbb{R}^{m \times n}$, where each column is a data point in dimension $m$, and a factorization rank $r$, an LRMF approximates $X$ as the product of two matrices of smaller inner dimension~$r$, $W \in \mathbb{R}^{m \times r}$ and $H \in \mathbb{R}^{r \times n}$, such that $X \approx WH$. 
The columns of $W$, denoted $W(:,k)$ for $k=1,\dots,r$, in dimension $m$, are called the basis vectors, and the entries of the $j$th column  of $H$, 
denoted $H(:,j)$ for $j=1,\dots,n$, indicate in which proportion each basis vector contributes to the corresponding data point, since $X(:,j) \approx WH(:,j)$.  

Without any constraints on the factors, $W$ and $H$, such a decomposition is highly non-unique. This has led researchers to use additional constraints, motivated by prior knowledge, to obtain uniqueness and \mbox{interpretability} of the factors. 
Two of the most widely used constraints are sparsity and \mbox{nonnegativity}. Such constrained LRMFs have had tremendous success in a wide variety of applications; see, e.g.,  \cite{lee1999learning, d2004direct, zou2006sparse, gribonval2010dictionary, fu2019nonnegative, gillis2020nonnegative}, and the references therein.

Recently, surfing the wave of deep learning, LRMFs have been extended such that the original matrix $X$ is decomposed as the product of more than two factors, which is referred to as deep MF.  
More precisely, $L$ layers of decomposition are applied on the matrix $X$ such that $X \approx W_LH_L H_{L-1} \dots H_1$. This is equivalent to considering a hierarchical decomposition of $X$ as follows:
\begin{align}
\label{eq:DeepMF_hier}
\begin{split}
     X &\approx W_1H_1, \\
    W_1 &\approx W_2H_2, \\ 
    &\mathrel{\makebox[\widthof{=}]{\vdots}} \\
    W_{L-1} &\approx W_L H_L. 
\end{split}
\end{align}
At each layer, $W_l \in \mathbb{R}^{m \times r_l}$  
and $H_l \in \mathbb{R}^{r_l \times r_{l-1}}$ for $l=1,\dots, L$ with $r_0=n$. 
The dimension $r_l$ of each layer is the rank of the factorization at layer $l$. When the factorization is performed according to the scheme of \eqref{eq:DeepMF_hier}, the ranks are generally chosen in decreasing order, that is $r_1 \geq r_2 \geq \dots \geq r_L$. This can be easily understood: the first levels of decomposition extract \mbox{low-leve bl} features, closely related to data while the last ones consist in a few key \mbox{high-level} features. Also, increasing ranks would lead to trivial factorizations. In fact, if $r_l \leq r_{l+1}$ for some $l$, $W_l = W_{l+1} H_{l+1} = (W_l \, 0) \binom{I_{r_l}}{0}$ where $0$ is the matrix of zeros of appropriate dimension, would be a feasible factorization that does not bring any informative feature.   

Deep MFs have the power to decompose hierarchically an input dataset, with different levels of interpretation at each layer, and are currently used in many applications, such as \revise{hyperspectral unmixing~\cite{feng2018hyperspectral}}, the extraction of facial features~\cite{cichocki2007multilayer, zhao2019deep, zhao2022progressive}, \revise{recommender systems \cite{wan2020deep}}, and multi-view  clustering~\cite{zhao2017multi, huang2020auto, luong2022multi} among others.

As for single-layer factorizations, additional constraints are typically assumed on the factors $W_l$'s and $H_l$'s. 
Indeed, without any constraint on the factors of deep MF, deep MFs degenerate into (overparametrized) classical MFs: 
the product of the factors $H_l$'s could be replaced by a single matrix whose rank is less than or equal to the minimum of the $r_l$'s. Hence, the LRMFs with constraints, such as sparsity and nonnegativity, have been extended to a multilayer frawework; see~\cite{de2021survey, chen2022survey} and the references therein for details. \revise{Similar to deep MF, deep concept factorization \cite{zhang2020deep} is an extension of archetypal analysis to several layers.}

A crucial question that has not been discussed thoroughly yet is the choice of the loss function used to assess the quality of the factorization. 
The loss functions proposed so far in the literature are not consistent because different losses are optimized at different layers; see the discussion in Section~\ref{sec:meth} for more details.  
Therefore, in this paper, we propose a general framework with appropriate loss functions to solve efficiently deep MFs with general constraints on the factors. Especially, in the numerical experiments, we illustrate the performance of our framework compared to the state of the art.

The remainder of the paper is organized as follows. In Section~\ref{sec:meth}, we describe the motivations behind our work regarding the state of the art. Then, in Section~\ref{sec:lossfunc}, we propose two new loss functions and discuss why they are more meaningful and consistent compared to previous works. 
Based on these new loss functions, we present in Section~\ref{sec:framework} a generic optimization framework relying on an extrapolated projected gradient method in order to tackle deep MFs with various constraints and regularizations. 
In Section~\ref{sec:synthe}, we present results on synthetic data, and in Section~\ref{subsec:real}, we compare the models on real hyperspectral and facial images, before concluding in Section~\ref{sec:ccl}.

\section{State of the art and motivations of the work}
\label{sec:meth}

The first factorization model with several layers was proposed by Cichocki et al.~\cite{cichocki2007multilayer}, and is usually dubbed ``multilayer matrix factorization''. 
It aims at minimizing $\|W_{l-1}-W_lH_l\|_F^2$, with $W_0=X$ successively for each layer $l=1,\dots,L$. In other words, each factorization of~\eqref{eq:DeepMF_hier} is performed one at a time, from the first to the last level, and the model is merely a succession of standard one-layer MFs. This hierarchical decomposition is sketched in Algorithm~\ref{algo1}, in which $\mathcal{W}_l$ and $\mathcal{H}_l$ denote the feasible sets respectively for the factors $W_l$'s and $H_l$'s, $l=1,\dots,L$, regardless of the specific constraints that apply (e.g., nonnegativity or sparsity). For each layer, a classical two block coordinate descent~(BCD) is performed. It updates $W_l$ and $H_l$ alternatively until some stopping criterion, such as a maximum number of iterations or an insufficient decrease of the loss function between two consecutive iterations, is reached.  Then, the algorithm moves to the next layer, which is in turn factorized and so on until the last layer ($l=L$).  At lines~\ref{algogo2:line1} and~\ref{algogo2:line2}, \textit{arg~reduce} denotes any algorithm that leads to a decrease of the corresponding loss function with the constraints enforced on  $H_l$'s and $W_l$'s.  
\setlength{\textfloatsep}{0pt}
    \begin{algorithm} [H]
        \caption{Multilayer matrix factorization~\cite{cichocki2006multilayer}} \label{algo1} 
 \begin{algorithmic}[1]
 \renewcommand{\algorithmicrequire}{\textbf{Input:}}
  \REQUIRE Nonnegative data matrix X, number of layers $L$,   inner ranks $r_l$'s and feasible sets $\mathcal{W}_l$ and $\mathcal{H}_l$ for $l=1,\dots,L$. 
 \renewcommand{\algorithmicrequire}{\textbf{Output:}}
  \REQUIRE Matrices 
  $W_1,\dots, W_L$ and  $H_1,\dots,H_L$.
   \STATE $W_0=X$
  \FOR {$l = 1, \dots, L$}
  \STATE Initialize $W_l^{(0)}$ and $H_l^{(0)}$
  \FOR {$k = 1, \dots$}
    \STATE $H_l^{(k)}=\underset{H \in \mathcal{H}_l}\argreduce \|W_{l-1}-W_l^{(k-1)}H\|_F^2$ \label{algogo2:line1}\\
    \STATE $W_l^{(k)}=\underset{W \in \mathcal{W}_l} \argreduce \|W_{l-1}-WH_l^{(k)}\|_F^2$\label{algogo2:line2}\\
  \ENDFOR
    \ENDFOR

 \end{algorithmic}
 \end{algorithm}
A pitfall of the approach of Cichocki et al.\ is that the factors of the last layers do not have any influence on those of the first ones since the first layers are factorized before the last ones. To remedy this, an iterative procedure was suggested by Trigeorgis et al.~\cite{trigeorgis2016deep}, called ``deep MF''. The initialization of the factors is performed through Algorithm~\ref{algo1} and is followed by iterative updates until some stopping criterion is met, as described in Algorithm~\ref{algo2}.
More precisely, once all the factors $W_l$'s and $H_l$'s have been updated once through sequential factorizations as in multilayer MF, several rounds of updates (starting with the first layer) are performed, taking into account the previous updates of the factors of the other layers. For this purpose, a global loss function was proposed by Trigeorgis et al.~\cite[Equation~(10)]{trigeorgis2016deep}, namely 
\begin{equation} 
   \mathcal{L}_0(H_1, H_2,\dots, H_L; W_L)= \|X-W_LH_L \dots H_2 H_1\|_F^2. 
    \label{eq:lossF}
\end{equation}
It is the squared Frobenius norm of the difference between the original matrix~$X$ and the approximation obtained by unfolding the $L$ layers of~\eqref{eq:DeepMF_hier}. This loss function~\eqref{eq:lossF} was reused by most of the papers in  the deep MF literature. 
\setlength{\textfloatsep}{0pt}
    \begin{algorithm} [H]
        \caption{Deep  MF with general constraints~\cite{trigeorgis2016deep}}  \label{algo2}
     
 \begin{algorithmic}[1]
 \renewcommand{\algorithmicrequire}{\textbf{Input:}}
  \REQUIRE Data matrix X, number of layers $L$, inner ranks $r_l$'s, \\ feasible sets $\mathcal{W}_l$ and $\mathcal{H}_l$ for $l=1,\dots,L$
 \renewcommand{\algorithmicrequire}{\textbf{Output:}}
  \REQUIRE Matrices 
  $W_1,\dots, W_L$ and  $H_1,\dots,H_L$

 \STATE Compute initial matrices $W_l^{(0)}$ and $H_l^{(0)}$ for all $l$ through a sequential decomposition of $X$ (for example Algorithm~\ref{algo1})

  \FOR {$k=1,\dots$}
  \FOR {$l = 1, \dots, L$}
   \STATE $A_l^{(k)}=\left\{
    \begin{array}{ll}
        W_L^{(k-1)} & \mbox{if } l=L \\
        W_{l+1}^{(k-1)}H_{l+1}^{(k-1)}  & \mbox{otherwise}
    \end{array}
\right. \label{algo2:my_line}$
\STATE $B_l^{(k)}=H_{l-1}^{(k-1)}\dots H_1^{(k-1)}$
  \STATE $H_l^{(k)}=\underset{H \in \mathcal{H}_l} \argreduce \|X-A_l^{(k)}HB_l^{(k)}\|_F^2$ \label{algo2:line1}\\
  \STATE $W_l^{(k)}=\underset{W \in \mathcal{W}_l}\argreduce \|X-WH_l^{(k)} B_l^{(k)}\|_F^2$ \label{algo2:line2}\\
  
  \ENDFOR

    \ENDFOR
 
 \end{algorithmic}
 \end{algorithm}
To understand more clearly how Algorithm~\ref{algo2} works, let us consider the simple case where $L=3$. In Table~\ref{tab:table1}, we report the loss function minimized at each stage of Algorithm~\ref{algo2}, regarding the updates of lines~\ref{algo2:line1} and~\ref{algo2:line2} with the updated factor in bold, the others being fixed. 
\begin{table}[!h]
\centering
\begin{tabular}{|c | c | c |} 
 \hline
 Layer $l$ & Update of $H_l$ & Update of $W_l$ \\
 \hline\hline
 1 & $\|X-W_2H_2\mathbf{H_1}\|_F^2$ & $\|X-\mathbf{W_1}H_1\|_F^2$\\
 2 & $\|X-W_3H_3\mathbf{H_2}H_1\|_F^2$  & $\|X-\mathbf{W_2}H_2H_1\|_F^2$\\
 3 & $\|X-W_3\mathbf{H_3}H_2H_1\|_F^2$ & $\|X-\mathbf{W_3}H_3H_2H_1\|_F^2$\\
 \hline
 \end{tabular}
    \caption{Loss functions minimized at each step of Algorithm~\ref{algo2} for $L=3$.}
    \label{tab:table1}
\end{table}

Table~\ref{tab:table1} shows that Algorithm~\ref{algo2} minimizes three different loss functions depending on which factor matrix is updated. More precisely, only the updates of the last layer ($l=3$) and the one of $H_2$ are performed according to the "global" loss function claimed in~\eqref{eq:lossF}. For $L>3$, the situation is even worse, hence Algorithm~\ref{algo2} does not appear to be coherent since different loss functions are minimized along the factors updates.  Consequently, convergence guarantees on this global loss function cannot be derived from such an optimization framework; see Fig.~\ref{fig:err1} and~\ref{fig:err2} in Section~\ref{subsubsec:faces} that shows that Algorithm~\ref{algo2} does not converge on a numerical example.

Note that one cannot simply minimize $\mathcal{L}_0$ in~\eqref{eq:lossF} at every layer to compute $(W_L,H_L,\dots,H_1)$. In fact, assuming the ranks are decreasing, that is, $r_{l+1} \leq r_{l}$ for all $l$ (this is the most reasonable setting; see Section~\ref{sec:intro}), any solution $(H_1, \dots, H_L; W_L)$ can be transformed into a degenerate solution $(H^{(*)}_1, \dots, H^{(*)}_L; W^{(*)}_L)$ with the same loss function, and with the following form 
\[
H^{(*)}_l\text{$=$}\left( 
\begin{array}{cc}
I_{r_L}                        & 0_{r_L \times (r_{l-1} - r_{L})}  \\
0_{(r_l - r_L) \times r_{L}}   & 0_{(r_l - r_L) \times (r_{l-1} - r_{L})}
\end{array} 
\right) 
\text{ for } l\text{$=$}2,\dots,L, 
\] 
where $I_{r_l}$ is the identity matrix of dimension $r_l$, and $0_{m \times n}$ is the $m$-by-$n$ zero matrix, $H^{(*)}_1=H_L...H_1$ and $W^{(*)}_L=W_L$.  
The reason is that $W_L H_L \dots H_1$ has rank at most $\min_l r_l = r_L$, and with the choice above, we have 

\begin{equation*}
    W^{(*)}_1 = W^{(*)}_L H^{(*)}_L \dots H^{(*)}_{2} = [W_L \, 0_{m \times (r_1-r_L)}] \in \mathbb{R}^{m \times r_1}
\end{equation*}
 such that $\text{rank}(W^{(*)}_1) \leq r_L$. This means that deep MF would simply reduce to an overparametrized LRMF~\cite{arora2019implicit}. Therefore, deep MFs models need to properly balance the importance of each layer to compute useful hierarchical decompositions.

Most of the recent works only rely on the loss given by~\eqref{eq:lossF} and follow Algorithm~\ref{algo2}. However, for the reasons mentioned above, this loss function and the use of Algorithm~\ref{algo2} 
are not consistent across layers, but this has not been much discussed in the literature yet. 
Besides, we show in Section~\ref{sec:exp} that this framework is not able to retrieve the ground truth basis vectors in practice. In fact, the loss function~\eqref{eq:lossF} only minimizes the error at the last layer of factorization but does not control the accuracy of the factorizations of the previous layers, which yet may be crucial regarding the applications of deep MF. 

To alleviate this, we propose in Section~\ref{sec:lossfunc} two new loss functions that are consistent, that is, are guaranteed to diminish after each update and reflect the balance between the errors due to each layer of decomposition.

 \section{Consistent loss functions for deep MFs} \label{sec:lossfunc}
 
 As explained in the previous section, the framework proposed by Trigeorgis et al.~\cite{trigeorgis2016deep} is inconsistent.
 In this section, we propose two global loss functions that can be used to optimize any of the factors in deep MF. 
 Hence, it is straightforward to derive meaningful update rules that ensure the decrease of the global loss function after the update of any factor. %, and \revise{provide convergence guarantees}.  

\subsection{Layer-centric loss function}  \label{subsec:loss1}

The first loss function proposed consists of a weighted sum of the errors caused by each layer of decomposition, that is, by the \mbox{layer-wise} factorizations:
\begin{multline}
   \hspace{-3mm} \mathcal{L}_1(H_1, H_2,\dots, H_L; W_1, W_2,\dots, W_L)=\frac{1}{2} \Big(\|X-W_1H_1 \|_F^2 \\ + \lambda_1 \|W_1-W_2H_2\|_F^2 + \dots + \lambda_{L-1} \|W_{L-1}-W_LH_L\|_F^2\Big).
    \label{eq:loss1}
\end{multline}
This loss function is quite intuitive, with each term corresponding to a \mbox{layer-wise error} as the factorizations unfold. In fact, this can be seen as a globalization of the model of Cichocki et al.~\cite{cichocki2006multilayer}: 
instead of trying to minimize the errors of each layer-wise factorization sequentially, all of them are aggregated within a global weighted loss function.

As $l$ increases and the ranks decrease (recall $r_l < r_{l-1}$ for all $l$), the computational cost to evaluate each term in~\eqref{eq:loss1} decreases. More precisely, the $l$-th term requires $m r_l r_{l-1}$ elementary operations to be computed hence the computational cost of evaluating~\eqref{eq:loss1} is $\mathcal{O} (Lmnr_1)$. The loss of the first layer, which corresponds to a standard factorization of the input matrix, is the most expensive to compute.

\subsection{Data-centric loss function} \label{subsec:loss2}

The second loss function considers the errors between $X$ and its successive approximations of ranks $r_l$'s: 
\begin{multline}
    \hspace{-3mm} \mathcal{L}_2(H_1, H_2,\dots, H_L; W_1, W_2,\dots, W_L)=\frac{1}{2}\Big( \|X-W_1H_1 \|_F^2 \\ + \mu_1 \|X-W_2H_2H_1\|_F^2 + \dots + \mu_{L-1} \|X-W_LH_L \dots H_2 H_1\|_F^2\Big).
    \label{eq:loss2}
\end{multline}
 
While the loss function described in Section~\ref{subsec:loss1} focuses on layer-wise errors, this one is data-centric in the sense that it evaluates the errors between the data matrix $X$ and its successive low-rank approximations. An advantage of this loss function is that the parameters $\mu_l$'s ($l=1,\dots,L-1$) are easier to tune since all the terms are likely to have a similar order of magnitude. However, since the successive approximations involve an increasing number of matrix multiplications, this loss function and the associated update rules are slightly more computationally expensive. Indeed, the most computationally costly term is the last one, which requires $n(mr_L+r_Lr_{L-1}+\dots + r_2r_1)$ operations to be computed, which may become high if the number of layers is high and the ranks do not decrease rapidly.

 \section{General algorithmic framework} \label{sec:framework}

A general algorithm to minimize the two proposed loss functions \eqref{eq:loss1} and~\eqref{eq:loss2} under general constraints on $W_l$'s and $H_l$'s 
is given in Algorithm~\ref{algo3}. We denote $W_{\rightarrow l}$ the set of matrices $\{W_1, \dots, W_{l-1}\}$ for any $l=1,\dots, L$ and $W_{l \rightarrow}$ the set of matrices \{$W_{l+1}, \dots, W_L$\}, and similarly for the $H_l$'s. 
\setlength{\textfloatsep}{0pt}
    \begin{algorithm} [H]
        \caption{Framework to solve deep MF with general constraints and consistent global loss function}  \label{algo3}
     
 \begin{algorithmic}[1]
 \renewcommand{\algorithmicrequire}{\textbf{Input:}}
  \REQUIRE Data matrix X, number of layers $L$, inner ranks $r_l$'s, \\  feasible sets $\mathcal{W}_l$ and $\mathcal{H}_l$ for $l=1,\dots,L$, a global loss function $\mathcal{L}$ such as  $\mathcal{L}_1$ in~\eqref{eq:loss1} or $\mathcal{L}_2$ in~\eqref{eq:loss2}
 \renewcommand{\algorithmicrequire}{\textbf{Output:}}
  \REQUIRE Matrices 
  $W_1,\dots, W_L$ and  $H_1,\dots,H_L$

 \STATE Compute initial matrices $W_l^{(0)}$ and $H_l^{(0)}$ for all $l$ through a sequential decomposition of $X$ (for example Algorithm~\ref{algo1})

  \FOR {$k=1,\dots$}
  
 \FOR {$l = 1, \dots, L$}
 
  \STATE $H_l^{(k)}=\underset{H \in \mathcal{H}_l} \argreduce \; \mathcal{L} \left(W_{\rightarrow l}^{(k)},W_l^{(k-1)}, W_{l \rightarrow}^{(k-1)}; H_{\rightarrow l}^{(k)}, \bm{H}, H_{l \rightarrow}^{(k-1)}\right)$  \label{algo3:line1}\\
 
  \STATE $W_l^{(k)}=\underset{W \in \mathcal{W}_l}\argreduce \; \mathcal{L} \left(W_{\rightarrow l}^{(k)},\bm{W}, W_{l \rightarrow}^{(k-1)}; H_{\rightarrow l}^{(k)}, H_l^{(k)}, H_{l \rightarrow}^{(k-1)}\right)$  \label{algo3:line2}\\
  
\ENDFOR
 \ENDFOR
 \end{algorithmic}
 \end{algorithm}

Algorithm~\ref{algo3} consists in BCD over the factors of each layer. 
The subproblems in one factor matrix (in bold at lines~\ref{algo3:line1} and~\ref{algo3:line2}) can be solved by various well-known techniques. In particular, when the feasible set is convex, these subproblems are convex. 
\revise{Moreover, many standard optimization schemes can be applied to such formulations and directly provide convergence guarantees to stationary points, e.g.,   the proximal alternating linearized minimization (PALM)~\cite{bolte2014proximal} and the block successive minimization (BSUM)~\cite{razaviyayn2013unified} optimization schemes.}  
This general framework is also very flexible and we present some examples below, considering usual constraints on the factors of each layer, which will be illustrated in the experiments of Sections~\ref{sec:synthe} and~\ref{subsec:real}.

One possibility to implement the updates of the factor matrices in Algorithm~\ref{algo3}, that is, to solve the \textit{arg reduce} subproblems at lines~\ref{algo3:line1} and~\ref{algo3:line2}, is a fast projected gradient method (FPGM), which is the one considered in the remainder of this paper.  This method is easy to implement and scales relatively well (linearly with the size of the data).  The FPGM is a well-known first-order optimization framework to update a general matrix $M$ which can be any of the $W_l$'s or $H_l$'s, $l=1,\dots, L$, see~Algorithm~\ref{algo4}. As each subproblem is convex, we choose $\frac{1}{L}$ as the step size, with $L$ the Lipschitz constant, except for the update of $H_l$'s for the second loss function $\mathcal{L}_2$. Indeed, in this case, the Lipschitz constant is quite costly to compute, as the derivatives are obtained by a sum over the layers (see~\eqref{eq:derivl2h}) hence, we simply compute the stepsize through a backtracking line search. The extrapolation step is based on Nesterov acceleration~\cite{nesterov1983method} and a restart guarantees the decrease of the loss function. In other words, if the error increases, the extrapolation step is not taken, as in~\cite{o2015adaptive}. 
\revise{Moreover, such an algorithm is guaranteed to converge to stationary points of the corresponding optimization problem, under suitable assumptions (in particular, the convexity of the subproblems in each factor matrix, which is our case here as long at $\mathcal{L}$ is block-wise convex, and the feasible set is convex)~\cite{xu2013block, xu2017globally}.}

\setlength{\textfloatsep}{0pt}
    \begin{algorithm} [H]
        \caption{Restarted fast projected gradient method (FPGM)}  \label{algo4}
     
 \begin{algorithmic}[1]
 \renewcommand{\algorithmicrequire}{\textbf{Input:}}
  \REQUIRE Initial matrix $M^{(0)}$, feasible set $\mathcal{M}$, loss function $f(M)$, parameter $\alpha_1 \in (0,1)$ 
   \renewcommand{\algorithmicrequire}{\textbf{Output:}}
  \REQUIRE A matrix $M$ that decreases $f$, that is, $f(M) < f(M^{(0)})$
 \STATE Compute Lipschitz constant $L$ of $f$; $Y=M^{(0)}$
 
 \FOR {$k=1,\dots$}
    \STATE $M^{(k)}=\mathcal{P}_{\mathcal{M}} \left(Y-\frac{1}{L}\nabla f(Y)\right)$ \label{algo4:lineProj}
    \STATE $Y=M^{(k)}+\beta_k \left(M^{(k)}-M^{(k-1)}\right)$ with $\beta_k=\frac{\alpha_k (1-\alpha_k)}{\alpha_k^2+\alpha_{k+1}}$ and $\alpha_{k+1}=\frac{1}{2}\left(\sqrt{\alpha_k^4+4\alpha_k^2}-\alpha_k^2\right)$
    \IF {$f\left(M^{(k)}\right) > f\left(M^{(k-1)}\right)$} 
    \STATE $Y=M^{(k-1 )}$, $\alpha_{k+1} = \alpha_1$ \emph{\% Restart} \ENDIF
\ENDFOR

 \end{algorithmic}
 \end{algorithm}

Let us compute the gradients with respect to $W_l$'s and $H_l$'s ($l=1,\dots,L$) of the loss functions~\eqref{eq:loss1} and~\eqref{eq:loss2}, by introducing $W_0=X$, $\lambda_0=\mu_0=1$. For $\mathcal{L}_1$, we have:
\begin{equation}
   \frac{ \partial{\mathcal{L}_1}} {\partial{W_l}}=\lambda_{l-1}(W_lH_l-W_{l-1})H_l^T+\delta_l \lambda_l (W_l-W_{l+1}H_{l+1})
   \label{eq:derivl1W}
\end{equation}
where $\delta_l=0$ if $l=L$ and $1$ otherwise,
\begin{equation}
   \frac{ \partial{\mathcal{L}_1}} {\partial{H_l}}=\lambda_{l-1}W_l^T(W_lH_l-W_{l-1}).
    \label{eq:derivl1h}
\end{equation}
For $\mathcal{L}_2$, let us call for all $l$, $D_l=H_{l-1}\dots H_1$, $\Tilde{H}_l=H_l\dots H_1= H_l D_l$ and $C_l^{(k)}=W_kH_k\dots H_{l+1}$ for all $k \geq l$ (for $k=l$, $C_l^{(l)}=W_l$).  Then, the gradients are given by:
\begin{equation}
  \frac{ \partial{\mathcal{L}_2}} {\partial{W_l}}=\mu_{l-1}(W_l\Tilde{H}_l-X)\Tilde{H}_l^T,
   \label{eq:derivl2W}
\end{equation}
\begin{equation}
  \frac{ \partial{\mathcal{L}_2}} {\partial{H_l}}=\sum_{k=l}^L \mu_{k-1}C_l^{{(k)}^T}(C_l^{(k)}H_lD_l-X)D_l^T.
   \label{eq:derivl2h}
\end{equation}

We now \revise{implement} this FPGM-based framework \revise{for} various deep MF's models that include constraints on the factors, such as non-negativity in Section~\ref{subsec:SSNMF} and sparsity in Section~\ref{subsec:sparseDNMF}, or add a regularization term to the loss function, such as a volume penalization in Section~\ref{subsec:MVDNMF}.

\subsection{Nonnegative Deep MF} \label{subsec:SSNMF}

In deep nonnegative MF (deep NMF), the factors of the decomposition~\eqref{eq:DeepMF_hier} are constrained to be nonnegative, that is, $W_l \geq 0$, $H_l \geq 0$ for all $l=1,\dots,L$. The projection operator $\mathcal{P}$  at line~\ref{algo4:lineProj} of Algorithm~\ref{algo4} simply consists of the projection on the nonnegative orthant, that is, $\mathcal{P}(A)=\max(A,0)$.

Other constraints could be easily incorporated, as long as the projection onto the feasible set can be computed efficiently. For example, a very common additional constraint consists in enforcing the entries of every column of the $H_l$'s to sum to~$1$, that is $\|H_l(:,j)\|_1 = 1$ for all $j=1,\dots,r_{l-1}$, for any~$l$. This expresses that the coefficients of the linear combination of basis vectors corresponding to each data point sum to $1$, hence can be interpreted as proportions; see e.g.,~\cite{abdolali2020simplex}.

\subsection{Sparse deep MF} \label{subsec:sparseDNMF}

Sparse matrix factorizations consist in enforcing some factors of the decomposition to be sparse to foster their interpretability. Numerous ways of tackling sparsity in MFs have been proposed in the literature, including targeting a row-wise (or column-wise) $l_1$ norm for some factors~\cite{hoyer2004non}, adding a $l_1$ and/or $l_2$ norm penalty~\cite{kim2007sparse} to the loss function, dictionary learning~\cite{gribonval2015sparse} and sparse component analysis~\cite{georgiev2005sparse}, among others.

Recently, an efficient and fast method, referred to as grouped sparse projection (GSP)~\cite{ohib2019explicit}, was developed to avoid the drawback of the methods mentioned beforehand, that is, the tuning of many parameters. Especially in the context of deep MF, where the number of factors to update grows linearly with the number of layers, it may be convenient to limit the number of parameters. Hence, for a given factor, GSP aims to reach a target average sparsity of the whole matrix instead of each row/column separately, which confers much more flexibility to the sparsity pattern compared to standard approaches. 

In Algorithm~\ref{algo4}, it suffices to consider at line~\ref{algo4:lineProj} the grouped sparse projection given in Algorithm~$1$ of~\cite{ohib2019explicit}. However, the feasible set of GSP is not convex, hence the convergence is not guaranteed when using the FPGM.

\subsection{Minimum-volume deep NMF} \label{subsec:MVDNMF}

Minimum-volume one-layer NMF (minVolNMF) is a well-known NMF variant~\cite{miao2007endmember,chan2009convex,fu2016robust} that encourages the basis vectors, that is, the columns of $W$, to have a small volume. Intuitively, this boils down to trying to make them as close as possible to the data points, which enhances the interpretability of the decomposition. The minVolNMF model on which we will focus adds a penalty term for the volume to the reconstruction error in the loss function and is expressed as:
\begin{equation}
    \underset{\substack{W \in \mathbb{R}_+^{m  \times r} \text{, } W^T\bm{e}=\bm{e} \\ H \in \mathbb{R}_+^{r  \times n} }}{\min} 
    \frac{1}{2} \left(\|X-WH\|_F^2 + \kappa \log \det(W^TW + \delta I)\right)
\end{equation}
where $\kappa$ and $\delta$ are parameters fixed by the user, and $\bm{e}$ is the column vector of all ones. It is important to note that, while many models consider the intuitive constraint $H^T\bm{e}=\bm{e}$, more recent approaches showed that imposing the \mbox{column-stochasticity} of $W$, that is, $W^T\bm{e}=\bm{e}$, instead of $H$ leads to better results in many applications. This is due to the better conditioning of $W$ in this case, see the discussion in Section~$4.3.3$ of~\cite{gillis2020nonnegative} for more details.

To the best of our knowledge, minVolNMF has not been extended to the deep context yet. We extend the approach of~\cite{fu2016robust} by incorporating a volume contribution at every layer to~\eqref{eq:loss1} and~\eqref{eq:loss2}. Hence, we add the following quantity to each term of both loss functions: for $l=1,2,\dots,L$, 
\begin{equation}
    \kappa_{l} \log \det(W_l^TW_l+\delta I_{r_l})
    \label{eq:lossMV1}
\end{equation}
while imposing column-stochasticity on every $W_l$. To solve the \mbox{minVolNMF} problem, a majorization-minimization (MM) framework is usually considered. This consists in minimizing a surrogate function, namely a strongly convex upper approximation of the loss function, see~\cite{fu2016robust} and~\cite{leplatminimum} for the details. 
The FPGM of Algorithm~\ref{algo4} can then be applied on this surrogate. 

The volume contribution implies an additional term in the gradients of both $\mathcal{L}_1$ and $\mathcal{L}_2$ w.r.t.\ $W_l$'s. More precisely, the term $\kappa_l W_lZ$, multiplied either by $\lambda_{l-1}$ or $\mu_{l-1}$ is added to~\eqref{eq:derivl1W} and~\eqref{eq:derivl2W} respectively, with $Z=( W_l^{{(*)}^T}W_l^{(*)} + \delta I_{r_l})^{-1}$ where $W_l^{(*)}$ denotes the last iteration of $W_l$ such that $Z$ is constant during a given update of $W_l$. 

A potential drawback of such an approach is the use of many \mbox{regularization} parameters, both for the weights of the linear combination of errors and the volume penalties at each layer. Indeed, in addition to the $L-1$ parameters $\lambda_l$'s of~\eqref{eq:loss1} or $\mu_l$'s of~\eqref{eq:loss2}, the user has to fix the values of the $L$ parameters $\kappa_l$'s involved in the volume regularization at each layer. In practice, the $\kappa_l$ for a given layer is set as follows. Given the initial error $err_l^{(0)}=\frac{1}{2}\|W_{l-1}^{(0)}-W_l^{(0)} H_l^{(0)}\|_F^2$ (for the \mbox{layer-centric} loss function \eqref{eq:loss1}) or $err_l^{(0)}=\frac{1}{2}\|X-W_l^{(0)} \tilde{H}_l^{(0)}\|_F^2$ (for the \mbox{data-centric} loss function \eqref{eq:loss2})  and a first guess $\tilde{\kappa}_l$, the final value $\kappa_l$ is given by $\kappa_l=\tilde{\kappa}_l\frac{err_l^{(0)}}{|\log \det(W_l^{{(0)}^{T}}W_l^{(0)} + \delta I_{r_l})|}$, such that the decomposition error and the volume term are of the same order of magnitude for a given layer.

\section{Numerical experiments}
\label{sec:exp}

In this section, we evaluate the models described in Section~\ref{sec:framework} on both synthetic (Section~\ref{sec:synthe}) and real (Section~\ref{subsec:real}) data. One drawback of the experiments carried on deep MF models is the lack of ground truth in datasets. Indeed, despite the hierarchical structure of many datasets, few of them have available ground truth at each layer. Moreover, it is generally hard to guess in advance how features of a given layer can be interpreted, especially when various constraints are applied on the factors. Even with synthetic data, the setting should be chosen carefully to guarantee easy interpretability of the factors.
For these reasons, we first present the results of our models on a simple yet meaningful toy example in dimension $m=3$. The low dimensionality offers the advantage to control the ground truth basis vectors and easily interpret the features at each layer. We then study the performance of our models for hyperspectral unmixing and the extraction of facial features to show their efficiency on real-world challenges.

A Matlab implementation of the framework described above, with all the experiments, is available on \url{https://bit.ly/flexDeepMF}. 
\subsection{Synthetic data}  
\label{sec:synthe}

We consider for all the experiments a 2-layers network (that is, $L=2$) in dimension $m=3$.  The ranks $r_l$'s are set to $r_1=6$, $r_2=3$, and the target basis matrices $W_1^{*}$ and $W_2^{*}$ are given by 
\[
W_1^{*}= \begin{pmatrix*}[c]
0.1 & 0.1 & 0.4 & 0.4 & 0.5 & 0.5 \\
0.4 & 0.5 & 0.1 & 0.5 & 0.1 & 0.4 \\
0.5 & 0.4 & 0.5 & 0.1 & 0.4 & 0.1
\end{pmatrix*}=W_2^{*}H_2^{*},\]
where $W_2^{*}=\begin{pmatrix*}[c]
1/2 & 0 & 1/2\\
0 & 1/2 & 1/2 \\
1/2 & 1/2 & 0
\end{pmatrix*}$ and $H_2^{*}=\begin{pmatrix*}[c]
0.2 & 0 & 0.8 & 0 & 0.8 & 0.2 \\
0.8 & 0.8 & 0.2 & 0.2 & 0 & 0  \\
0 & 0.2 & 0 & 0.8 &  0.2 &  0.8
\end{pmatrix*}$.\\ 

Each column of the matrix $H_1^{*}$ is generated according to a Dirichlet distribution of parameter $\alpha=0.05$. The data matrix $X$, made of $n=1000$~points, is therefore generated as $X=X^{*}+N$ where $X^{*}=W_1^{*}H_1^{*}$ and $N$~is additive Gaussian noise: $N=\epsilon \|X^{*}\|_F \frac{Y}{\|Y\|_F}$ with $Y \sim \mathcal{N}(0,1) $. 

\begin{figure*}
   \centering
    \includegraphics[scale=0.3]{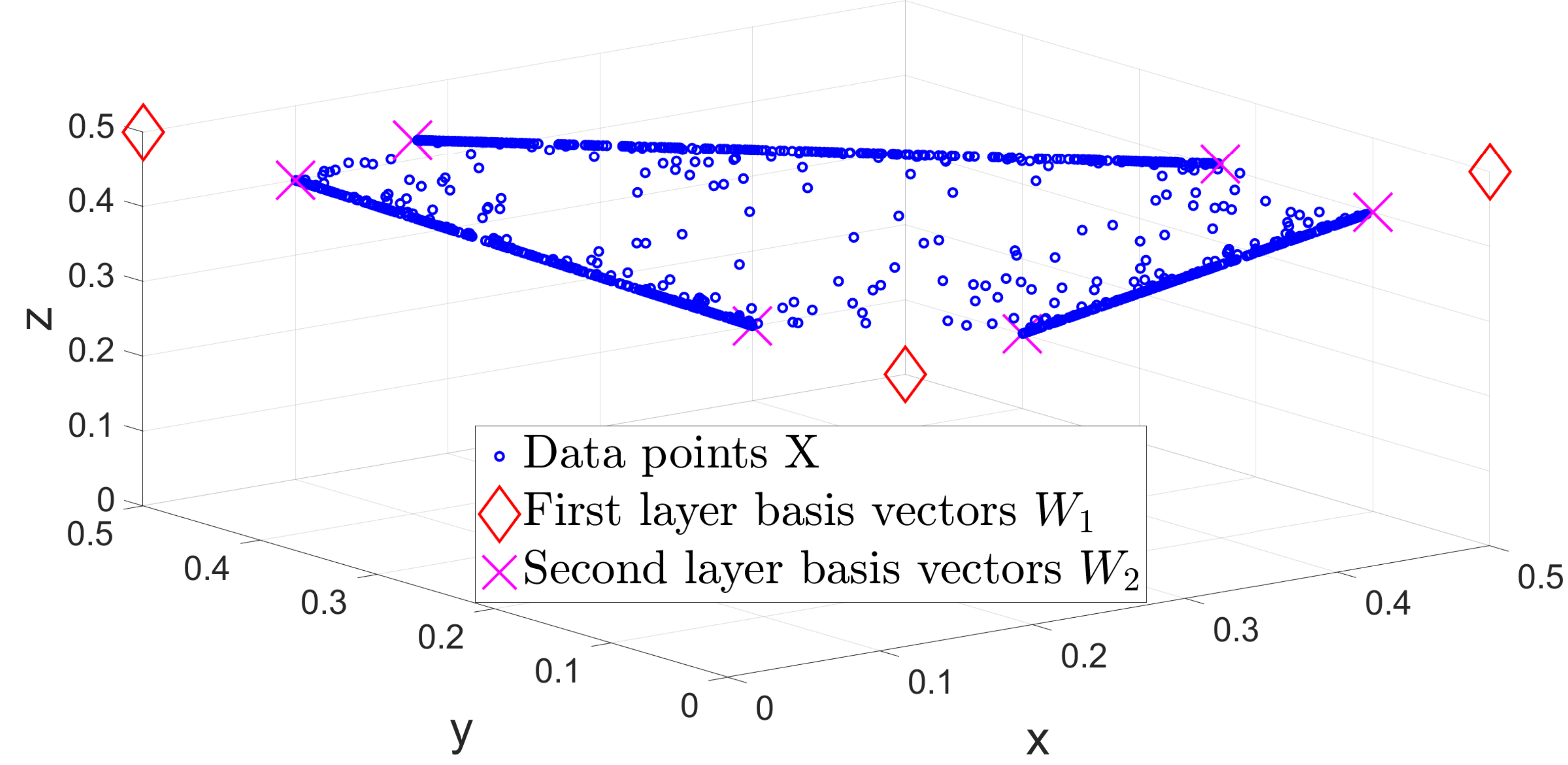}
    \caption{Setting of the synthetic data considered in this section, in the noiseless case.} 
    \label{fig:synthe_setting}
\end{figure*}

An example of a data set generated in that way in the noiseless case~($\epsilon=0$) is presented on Fig.~\ref{fig:synthe_setting}, with the ground truth basis vectors at both layers. In the following, we consider $10$ levels of noise: $\epsilon= 10^{-2}$, $2.51 \; 10^{-2}$, $6.31 \; 10^{-2}$, $9.49 \; 10^{-2}$, $1.267 \; 10^{-1}$, $1.585 \; 10^{-1}$, $2.384 \; 10^{-1}$, $3.182 \; 10^{-1}$, $3.981 \; 10^{-1}$, $1$.

We compare five models, namely: 
\begin{itemize}

\item Single-layer NMF of ranks $r_l$, $l=2,\dots,L$, 
 
    \item The sequential multilayer MF of \cite{cichocki2006multilayer}, as described in Algorithm~\ref{algo1}, dubbed MMF. At the first layer, the solution of \mbox{single-layer} NMF corresponds to the one of MMF, by construction. 
    
      \item The deep MF model from~\cite{trigeorgis2016deep}, see  Algorithm~\ref{algo2}, dubbed Tri-DMF. Although the updates of the original paper are performed with multiplicative updates (MU), we decided to solve the subproblems at lines~\ref{algo2:line1} and~\ref{algo2:line2} with FPGM since MU are known to be slow. 
    
    \item Deep MF with the layer-centric loss function \eqref{eq:loss1} solved with Algorithms~\ref{algo3} and~\ref{algo4}, dubbed LC-DMF,
    
    \item Deep MF with the data-centric loss function \eqref{eq:loss2} solved with Algorithms~\ref{algo3} and~\ref{algo4}, dubbed DC-DMF.

\end{itemize}
The global loss functions of these methods are not meaningfully comparable to each other due to, on the one hand, the absence of global loss function for multilayer MF and, on the other hand, the difference of magnitude between the terms of layer-centric and data-centric loss functions (see Section~\ref{sec:lossfunc}). Therefore, we report the average mean-removed spectral angle~(MRSA) between the corresponding expected and computed basis vectors, that is, the columns of $W_l^{*}$ and $W_l$ respectively at each layer $l$. Given any two vectors $a$ and $b$, their MRSA is defined as 
\begin{equation*}
    MRSA(a,b) = \frac{100}{\pi} \text{arcos} \Big(\frac{\langle a - \overline{a}, b - \overline{b}\rangle}{\Vert a - \overline{a}\Vert_{2}\Vert b - \overline{b}\Vert_{2}}\Big) \in \left[  0, 100\right]
\end{equation*}
where~$ \langle \cdot, \cdot\rangle $~indicates the scalar product of two vectors and~$\overline{\cdot}$~is the mean of a vector.

For all methods, the initial factors $W_l^{(0)}$ and $H_l^{(0)}$ are obtained by projecting onto the feasible sets the output of the successive nonnegative projection algorithm (SNPA)~\cite{gillis2014successive} applied on $W_{l-1}^{(0)}$, with $W_0^{(0)}=W_0=X$. In a nutshell, SNPA is a column subset selection algorithm often used as an initialization technique for NMF.   

\revise{
\begin{remark}[Compared algorithms and goal of our experiments]  
Our main goal in this paper is to show that our new proposed loss functions, \eqref{eq:loss1} and \eqref{eq:loss2}, are much more meaningful and lead to significantly better results in practice than the most widely used one, namely, that of Trigeorgis et al.~\cite{trigeorgis2016deep}, that is, Tri-DMF.  
In fact, to the best of our knowledge, most papers on deep MF rely on Tri-DMF. Their novelty is typically embedded into the regularizers and/or constraints added into their model to tackle specific applications: for example, 
a total variation regularizer for hyperspectral unmixing in~\cite{feng2018hyperspectral}, 
a regularizer to enhance good local characteristics for basis image extraction in~\cite{zhao2019deep}, 
or a so-called community regularization for recommendation in social networks in~\cite{wan2020deep}. 
 We do not compare to such more recent deep MF models because, to have a meaningful and fair comparison, we would need to adapt our model as well, adding proper  regularizers. Moreover, we would need to provide background on each particular application, which is out of the scope of this paper.  
 In summary, because our main goal is to provide theoretical and experimental evidence of the superiority of our proposed models, and the failure of the model of Trigeorgis et al.~\cite{trigeorgis2016deep}, we do not to focus on particular models designed for specific applications and try to be as generic as possible. 
 The regularizations and constraints we consider in this paper are very general  and apply to a wide range of applications.   
\end{remark}
}

\paragraph{\revise{MinVol deep NMF}}  We first consider the minVol deep NMF variant (minVol regularization together with non-negativity of the factors). The parameter $\delta$ is fixed to $\delta=0.1$. For the $\tilde{\kappa}_l$'s (see Section~\ref{subsec:MVDNMF}), we make a distinction between the first 4 levels of noise ($\epsilon < 0.1$) and the last 6 ($\epsilon > 0.1$), since minimizing the volume is more challenging when the noise increases. More precisely, we fix  $\tilde{\kappa}_1=10^{-3}$, $\tilde{\kappa}_2=10^{-2}$ for the four first levels and $\tilde{\kappa_1}=10^{-2}$, $\tilde{\kappa}_2=10^{-1}$ for the six last levels of noise, for all the compared methods. To compute the parameters $\lambda_l$'s, $l=1,...,L-1$ in \eqref{eq:loss1}, we proceed similarly to what is done for the minVol parameters $\kappa_l$'s, by always considering the first layer of decomposition as a baseline. More precisely, based on an initial guess $\tilde{\lambda}_l$, we set $\lambda_l=\tilde{\lambda}_l \frac{err_1^{(0)}}{err_{l+1}^{(0)}}$, where $err_k^{(0)}$ denotes the $k$-th layer error $\frac{1}{2}\|W_{k-1}-W_kH_k\|_F^2$ after the initialization. By doing so, the ratio between the $(l+1)$-th and the first term of \eqref{eq:loss1} is approximately equal to an arbitrary value $\tilde{\lambda}_l$, fixed by the user. In practice, we used $\tilde{\lambda_l}=10$ for all $l=1,\dots, L-1$. For the parameters $\mu_l$'s of the data-centric loss function, we fixed $\mu_l=1$ for all $l=1,\dots, L-1$ since all the reconstruction error terms are expected to be of the same order of magnitude. We use the same values of these parameters for the experiments on real data in Section~\ref{subsec:real}.

\begin{figure*}

    \centering
    \includegraphics[scale=0.3]{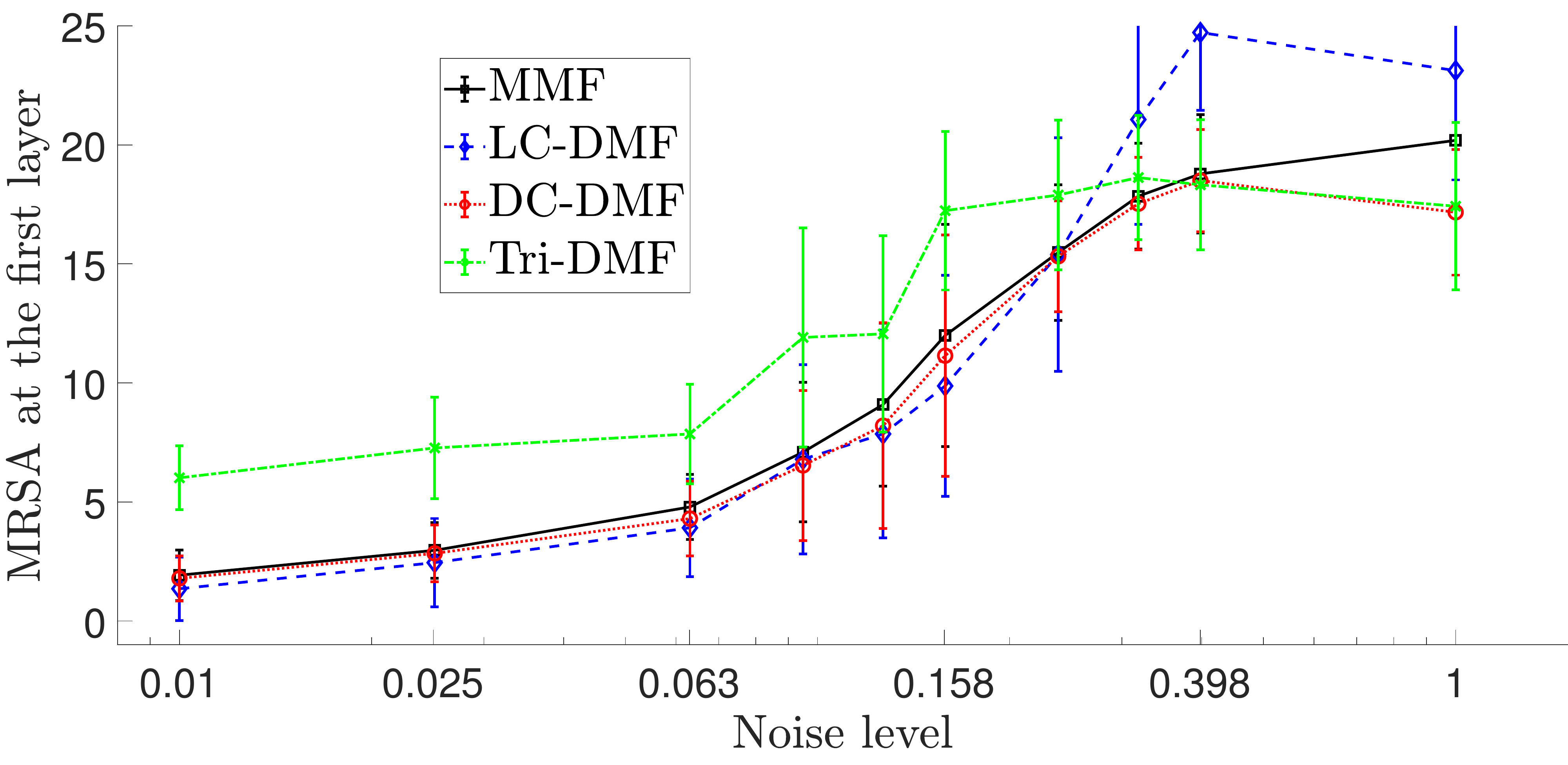}
    \caption{Comparison of the MRSA obtained at the first layer with minVol MMF, \mbox{LC-DMF}, \mbox{DC-DMF} and \mbox{Tri-DMF} on synthetic data in function of the noise level.} 
    \label{fig:MRSA1}
\end{figure*}

\begin{figure*}

    \centering
    \includegraphics[scale=0.3]{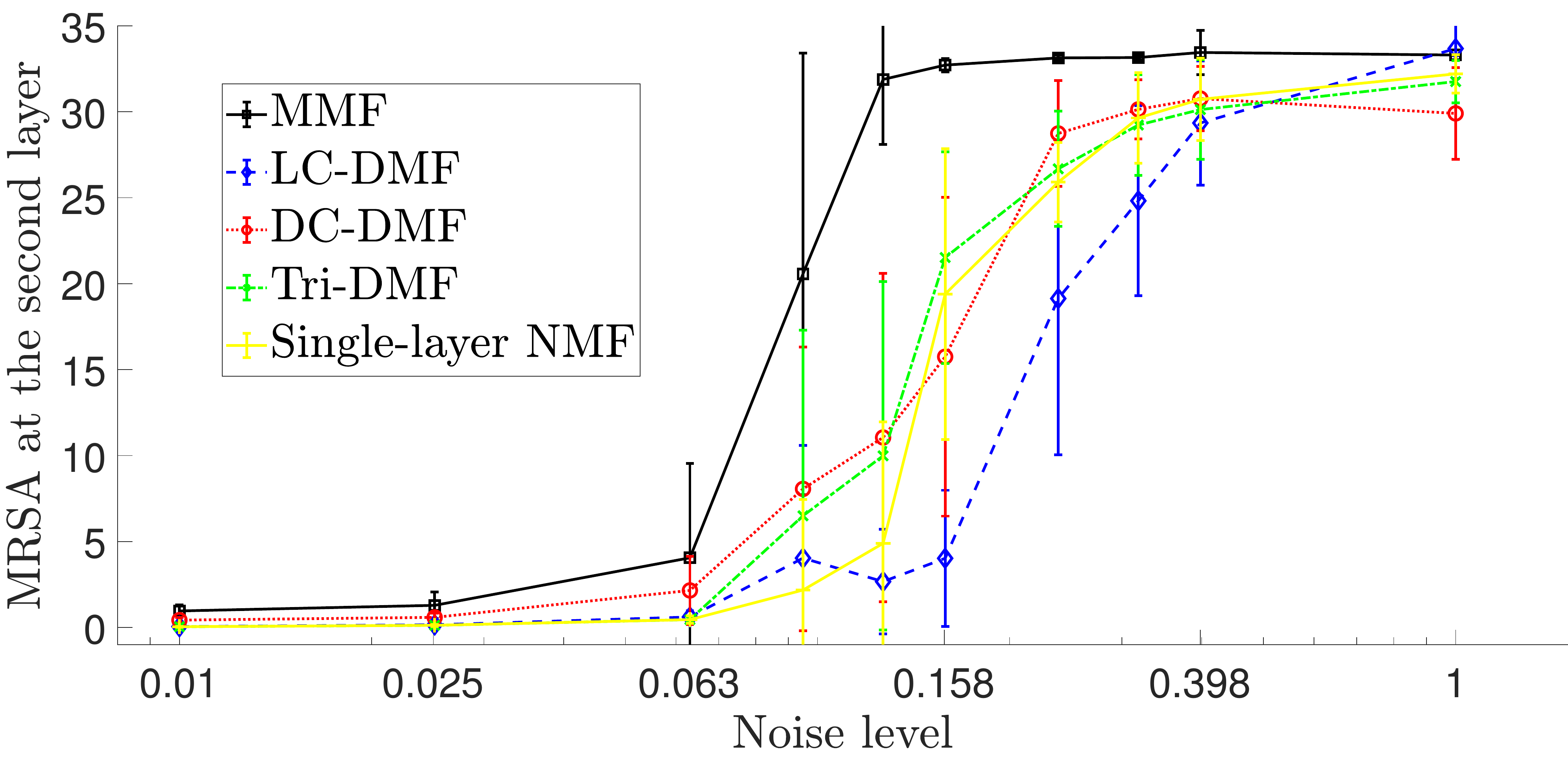}
    \caption{Comparison of the MRSA obtained at the second layer with minVol MMF, \mbox{LC-DMF}, \mbox{DC-DMF}, \mbox{Tri-DMF} and \mbox{single-layer} NMF on synthetic data in function of the noise level.}
    \label{fig:MRSA2}
\end{figure*}

 Fig.~\ref{fig:MRSA1}~and~\ref{fig:MRSA2} display the MRSA (average and standard deviation over 25 runs) of MMF, \mbox{LC-DMF}, \mbox{DC-DMF}, \mbox{Tri-DMF} and \mbox{single-layer} NMF (for the second layer only) in function of the noise level for the first and second layers, respectively. 
For both layers, LC-DMF produces the lowest MRSA, except for very high noise levels.   \mbox{DC-DMF} becomes more competitive as the noise increases. 
 On the contrary, especially at the second layer, MMF produces a higher MRSA than \mbox{DC-DMF} and \mbox{LC-DMF}. This confirms that a weighted loss function together with an iterative framework such as Algorithm~\ref{algo3} is more efficient than the purely sequential approach of Cichocki et al. 
On the other side, 
the approach of Trigeorgis et al.\ completely fails to retrieve the correct basis vectors at the first layer, with one or two predicted basis vectors located inside the convex hull of the others. Even at the second layer, \mbox{Tri-DMF} is not the best method although the corresponding loss function is designed to minimize the reconstruction error at the last layer. Finally, \mbox{single-layer} NMF also performs worse than \mbox{LC-DMF} at the second layer and, of course, does not allow to automatically bind the features of consecutive layers, which is undoubtedly the main added value of deep approaches.

\paragraph{\revise{Sparse deep MF}} Let us now investigate sparse deep MF. Due to the structure of the ground truth factors, we only considered sparsity on the factors of the second layer, fixing the target grouped sparsity of both $W_2$ and $H_2$ to $\frac{1}{3}$. The values of~the parameters $\lambda_l$'s and $\mu_l$'s are chosen in the same way as for the minVol variant.

\begin{figure*}
    \centering
    \includegraphics[scale=0.3]{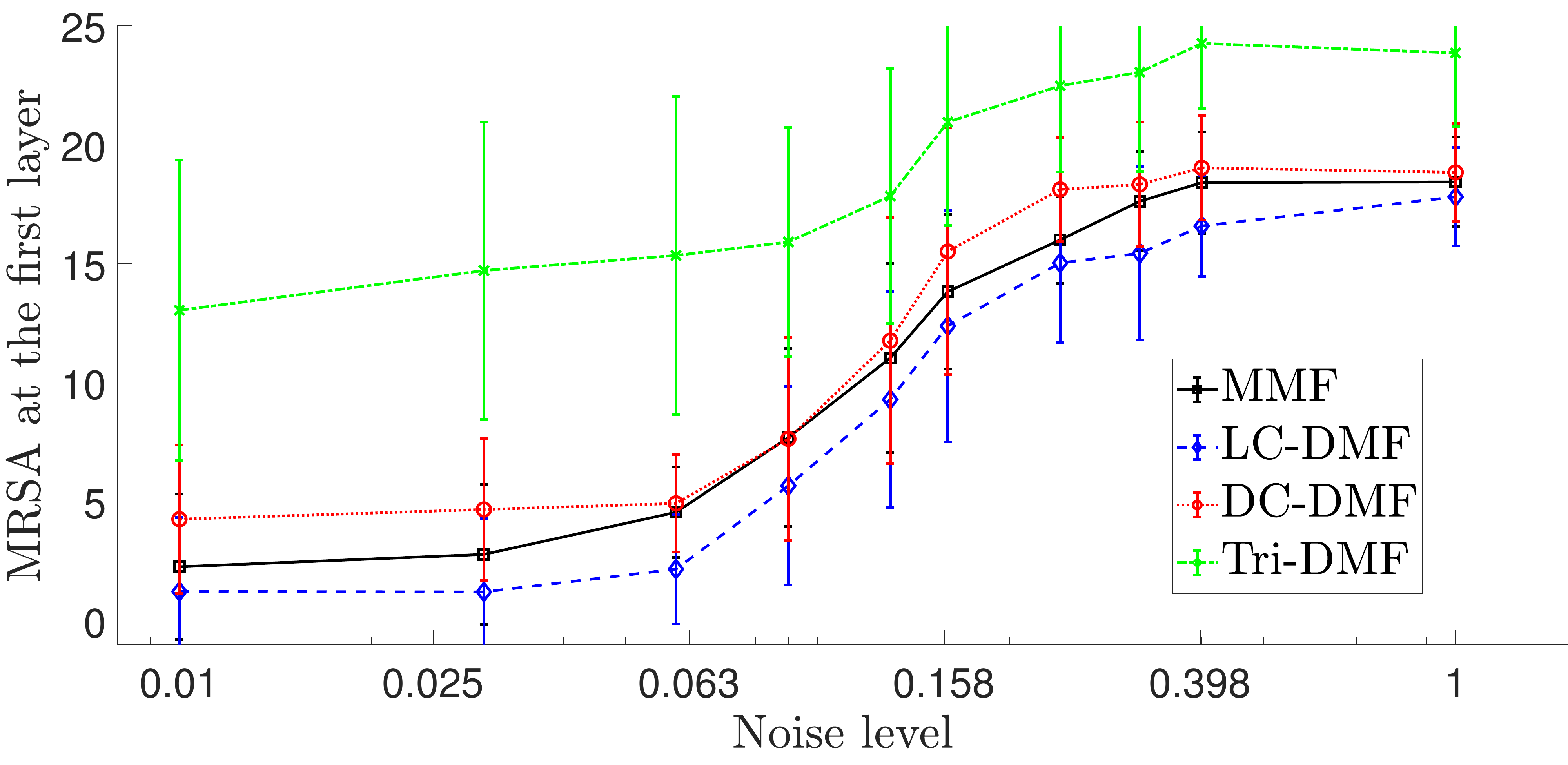}
    \caption{Comparison of the MRSA obtained at the first layer with grouped sparse MMF, \mbox{LC-DMF}, \mbox{DC-DMF} and \mbox{Tri-DMF}  on synthetic data in function of the noise level.}
    \label{fig:MRSA_bis1}
\end{figure*}

\begin{figure*}
    \centering
    \includegraphics[scale=0.3]{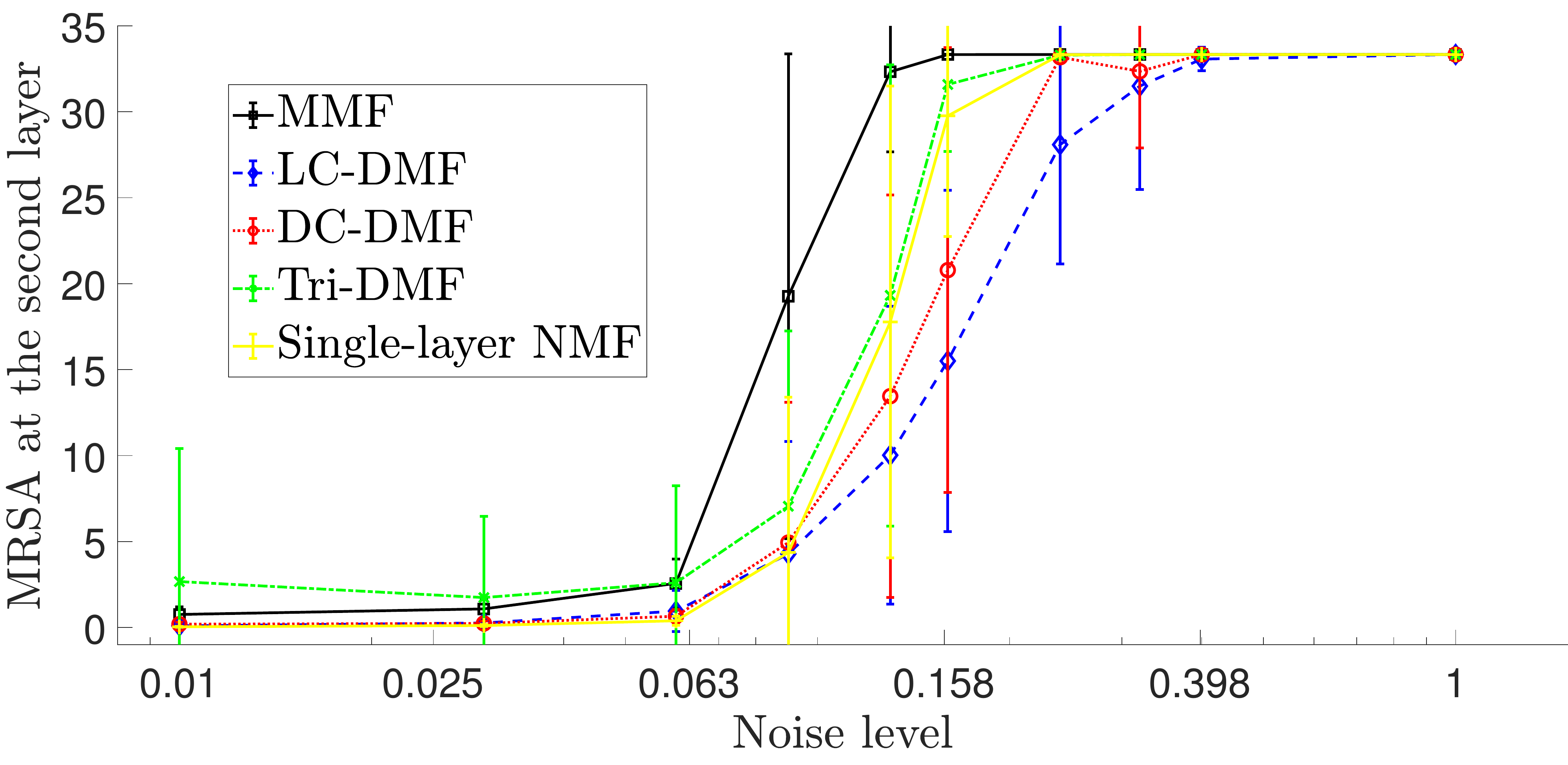}
    \caption{Comparison of the MRSA obtained at the second layer with grouped sparse MMF, LC-DMF, DC-DMF, Tri-DMF and \mbox{single-layer} NMF on synthetic data in function of the noise level.}
    \label{fig:MRSA_bis2}
\end{figure*}

 Fig.~\ref{fig:MRSA_bis1} and~\ref{fig:MRSA_bis2} display the MRSA of MMF, \mbox{LC-DMF}, \mbox{DC-DMF}, \mbox{Tri-DMF} and \mbox{single-layer} NMF in function of the noise level for the first and second layers, respectively. 
 The conclusions are similar to those of minVol deep~MF: 
 \mbox{LC-DMF} performs better in terms of MRSA than \mbox{DC-DMF} and MMF, and \mbox{Tri-DMF}  completely fails to recover the basis vectors at the first layer. Moreover, at the second layer, \mbox{single-layer} NMF performs worse than weighted deep approaches such as \mbox{LC-DMF} and \mbox{DC-DMF}. This tends to confirm that \mbox{LC-DMF} is the most effective to recover the ground truth factors. 

\subsection{Real data} \label{subsec:real}

In this section, we present the performance of our framework on real data, namely hyperspectral images in Section~\ref{subsubsec:hyper} and faces in Section~\ref{subsubsec:faces}.
\medskip
\subsubsection{Hyperspectral unmixing}  
\label{subsubsec:hyper}

A hyperspectral image (HI) is characterized by the reflectance values of $n$ pixels in $m$ wavelength spectral bands and is generally represented by a matrix $X \in \mathbb{R}^{m \times n}$ where each column of $X$ is the spectral signature of one pixel. Hyperspectral unmixing (HU) aims to identify the spectral signatures of $r$ materials and under the linear mixing assumption, NMF has been widely used to solve HU~\cite{bioucas2012hyperspectral}. When deep NMF is applied, the materials are extracted in a hierarchical manner \cite{tong2017hyperspectral, de2021deep}.   

\begin{figure*}[!b]
\begin{center} 
\includegraphics[scale=0.6]{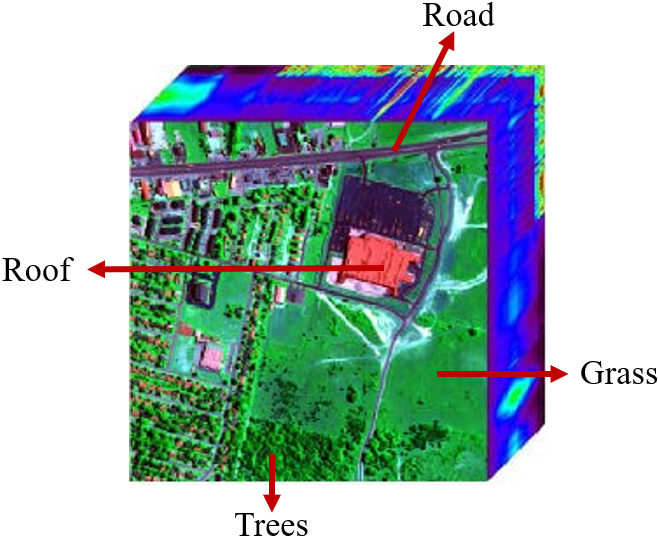}  
\caption{Urban hyperspectral image and its four main materials.  \label{fig:hyper_Urban}}
\end{center}
\end{figure*}

We apply minVol deep NMF on the HYDICE Urban HI, which is made of $n=307 \times 307$ pixels in $m=162$ spectral bands~\cite{zhu2017spectral}; see Fig.~\ref{fig:hyper_Urban}. 

We consider a $3$-layers network, with $r_1=6$, $r_2=4$ and $r_3=2$. To initialize the basis vectors of all layers, we use hierarchical clustering (HC)~\cite{gillis2015hierarchical} instead of SNPA. Indeed, SNPA is designed to extract extreme points of the dataset which may not be the most appropriate for the noisy, high-dimensional hyperspectral data. On the other hand, HC extracts clusters centroids. After several trials, it turns out that appropriate values for the minVol hyperparameters $\tilde{\kappa}_l$'s are $10^{-2}$ for all $l$. 

As the hyperspectral data are high-dimensional and tough to unmix, instead of simply initializing the factors of \mbox{LC-DMF}, \mbox{DC-DMF} and \mbox{Tri-DMF} with HC, we perform a few iterations of BCD after running HC at each layer, similarly to MMF (see Algorithm~\ref{algo1}), before moving to the initialization of the next layer. After this "augmented" initialization, the framework of 
Algorithm~\ref{algo3} is applied. More precisely, we run the same number $it=500$~iterations of MMF, \mbox{LC-DMF}, \mbox{DC-DMF} and \mbox{Tri-DMF}, among which, for the last three methods, $it_{in}=50$~iterations are devoted to improve the initial factors of all layers and the remaining consist in the iterative updates of Algorithm~\ref{algo3}. We also run single-layer NMF for $r=4$ and $r=2$ (at the first layer, it coïncides with MMF) to evaluate the efficiency of deep approaches compared to a shallow one.

\begin{figure*}
\hspace*{-0.15in}
    \centering
    \includegraphics[scale=0.5]{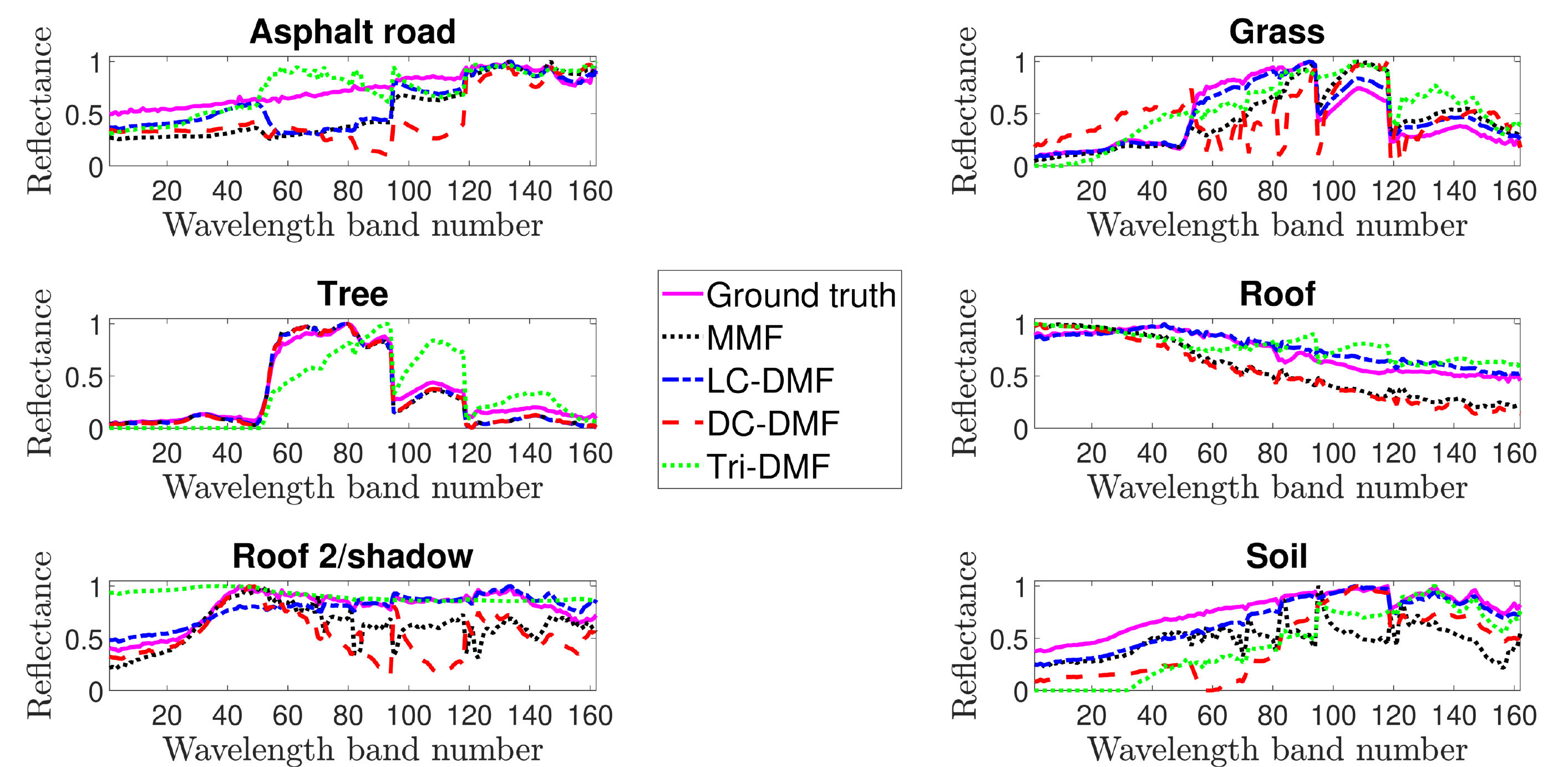}
    \caption{Endmembers extracted by MMF, LC-DMF, DC-DMF and Tri-DMF in the Urban image at the first layer ($r_1=6$), and the ground truth.}
    \label{fig:end6}
\end{figure*}

\begin{figure*}
\hspace*{-0.15in}
    \centering
    \includegraphics[scale=0.48]{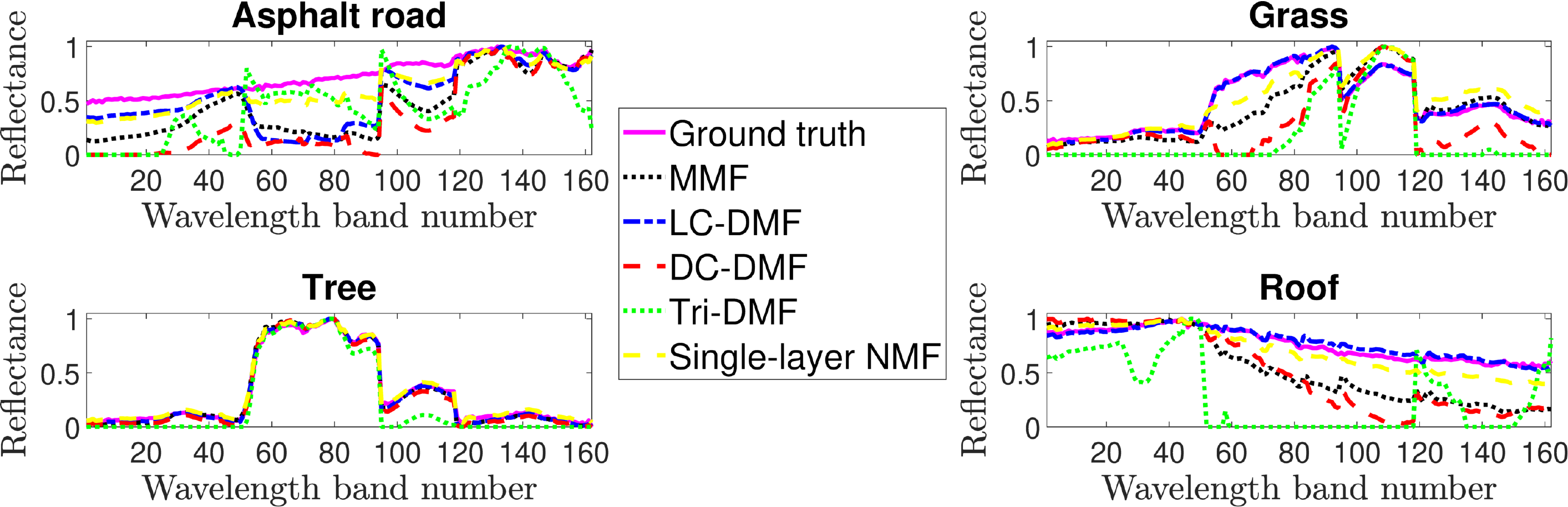}
    \caption{Endmembers extracted by MMF, LC-DMF, DC-DMF, Tri-DMF and single-layer NMF in the Urban image at the second layer ($r_2=4$), and the ground truth.} 
    \label{fig:end4}
\end{figure*}

On Fig.~\ref{fig:end6} and~\ref{fig:end4}, we plot the spectral signatures of the materials extracted by the pre-cited methods at the first and second layer respectively, that is, the columns of $W_1$ and $W_2$, and the ground truth from \cite{zhu2017spectral}, which is only available for $r=6$ and $r=4$ to the best of our knowledge. The MRSA's at both layers are presented in Table~\ref{tab:MRSA_table}. At both layers, \mbox{LC-DMF} outperforms the other deep methods, including \mbox{Tri-DMF} and MMF. Moreover, at the second layer, \mbox{LC-DMF} achieves a MRSA very close to the one of \mbox{single-layer} NMF. Therefore, as for the synthetic data, \mbox{LC-DMF} seems to be the best loss function to minimize when tackling deep MF.

\begin{table} [h]
\centering
\begin{tabular}{| c || c | c |}
 \hline
 Method & First layer ($r_1=6$) & Second layer ($r_2=4$) \\ [0.5ex] 
 \hline\hline
MMF & 16.98 & 12.35\\ 
 \hline
LC-DMF & $\bm{9.48}$ & $8.42$\\
 \hline
DC-DMF & 22.95 & 14.15\\
 \hline
Tri-DMF & 26.07 & 20.07 \\
 \hline
Single-layer NMF & 16.98 & $\bm{7.74}$\\ 
 \hline
\end{tabular}
\caption{MRSA of the compared methods at the first and second layer on the Urban hyperspectral image, with in bold the best value of each column.}
\label{tab:MRSA_table}
\end{table}

In ~\ref{subsec:appendA}, we present the abundance maps indicating the proportion of every material in each pixel at each layer, extracted by MMF, \mbox{LC-DMF}, \mbox{DC-DMF}, \mbox{Tri-DMF} and \mbox{single-layer} NMF on Fig.~\ref{fig:Abund_MMF}, \ref{fig:Abund_LC}, \ref{fig:Abund_DC}, \ref{fig:Abund_Tri}, \ref{fig:single4} and \ref{fig:single2} respectively.  In general, the six materials extracted at the first layer are asphalt road, grass, tree, two types of roof tops, and soil. Finally, at the last layer, two main categories of materials remain: on the one hand, grass and trees are mainly merged in a single "vegetal" cluster while on the other hand, the other materials are combined in a second "non-vegetal" cluster. On the figures, the arrows between materials extracted at consecutive layers indicate which materials of a given layer contribute to those of the next layer (that is, the arrows represent the non-zero entries of the corresponding $H_l$). When a material of the upper layer contributes to less than 10\% to a material of the lower layer, the arrow is discarded. We observe that the hierarchy of materials extracted by \mbox{LC-DMF} is rather sparse, as each material is obtained by a combination of only a few materials of the previous layer. On the opposite, the decomposition of \mbox{Tri-DMF} is hardly interpretable since all materials contribute almost equally to those of the next layer.

\medskip

\subsubsection{Facial features extraction}   
\label{subsubsec:faces}

Deep MF has also been shown efficient to extract facial features hierarchically~\cite{cichocki2006multilayer,trigeorgis2016deep}. 
The CBCL image\footnote{\url{http://www.ai.mit.edu/courses/6.899/lectures/faces.tar.gz}} is made of 2429 grey-scale images of~$19\times 19$~pixels representing faces of different people exhibiting various expressions.

\begin{figure} [!b]
  \centering
  \subfloat[][]{\includegraphics[width=0.29\textwidth]{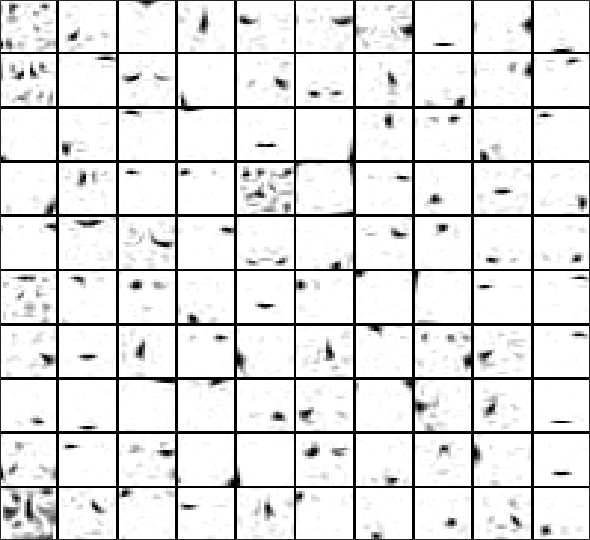}
  \label{fig:F1}}
  \hspace{4mm}
   \subfloat[][]{\includegraphics[width=0.29\textwidth]{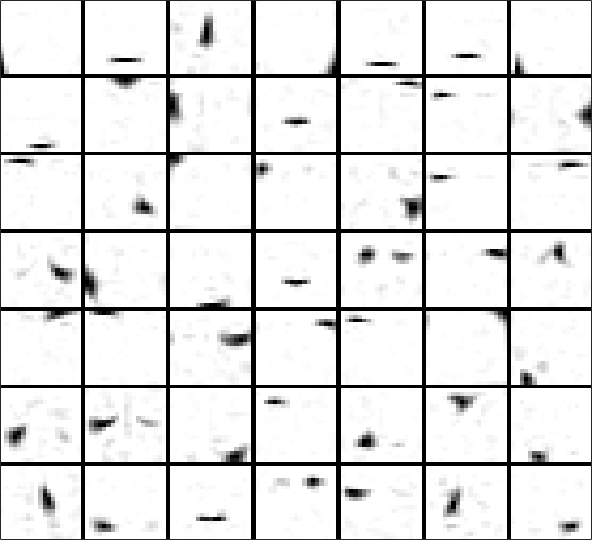}
  \label{fig:F2}}
  \hspace{4mm}
   \subfloat[][]{\includegraphics[width=0.29\textwidth]{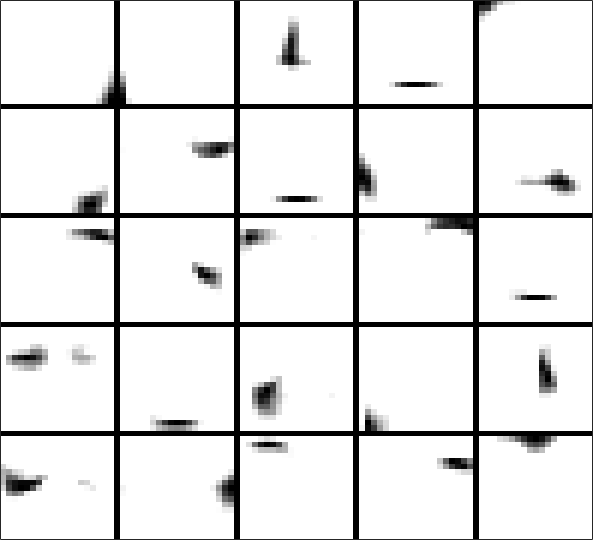}
  \label{fig:F3}}    

\caption{Features extracted by MMF on the CBCL face data set, with $L=3$, $r_1=100$, $r_2=49$, and $r_3=25$. Each image contains the features extracted at a layer: \protect\subref{fig:F1} first layer $W_1$, 
\protect\subref{fig:F2} second layer $W_2$, and \protect\subref{fig:F3} third layer $W_3$.
}
\label{fig:faces1}
\end{figure}

For this application, we consider grouped sparsity constraints on the factors. We set $L=3$, $r_1=100$, $r_2=49$ and $r_3=25$, and perform the initialisation of the factors with SNPA. We run MMF, \mbox{LC-DMF}, \mbox{DC-DMF} and \mbox{Tri-DMF} with a grouped sparsity level, that is, the average Hoyer sparsity of the columns, of $70\%$ on $W_1$, $80\%$ on $W_2$ and $85\%$ on $W_3$. Given any column vector $x$, its Hoyer sparsity~\cite{hoyer2004non} is defined as $sp(x)=\frac{\sqrt{n}-\frac{\|x\|_1}{\|x\|_2}}{\sqrt{n}-1}$. 
In comparison, the average Hoyer sparsity of $W_1$, $W_2$ and $W_3$ obtained with MMF without sparsity constraints are respectively $71.97\%$, $79.46\%$ and $76.64\%$.  The features extracted at all layers by MMF are displayed on Fig.~\ref{fig:faces1}. It shows the hierarchical decomposition of the data set by MMF: the larger facial features extracted at the first levels are made of smaller ones extracted at the deeper levels, such as eyes, mouths, and eyebrows. Sparsity allows the features, especially at the last layer, to contain only a few activated pixels. 
The features extracted by the other methods, that is, \mbox{LC-DMF}, \mbox{DC-DMF}, \mbox{Tri-DMF} and single-layer NMF are displayed in \ref{subsec:appendB} on Fig.~\ref{fig:faces2}, \ref{fig:faces3}, \ref{fig:faces4} and~\ref{fig:faces5}, respectively.

\begin{figure*}
  \centering
   \subfloat[][]{\includegraphics[width=0.7\textwidth]{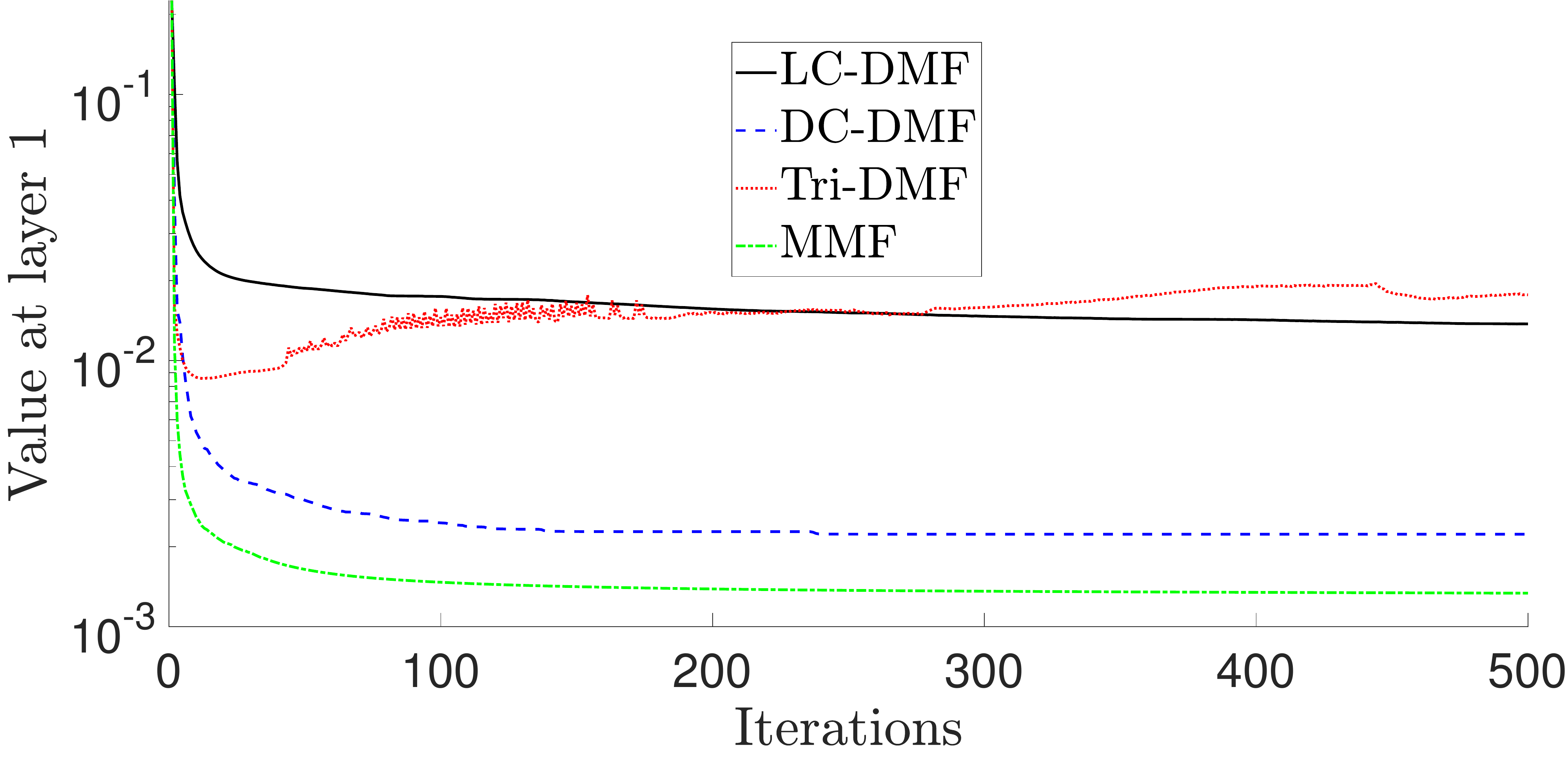}
  \label{fig:2A}}
  \hspace{3mm}
  \subfloat[][]{\includegraphics[width=0.7\textwidth]{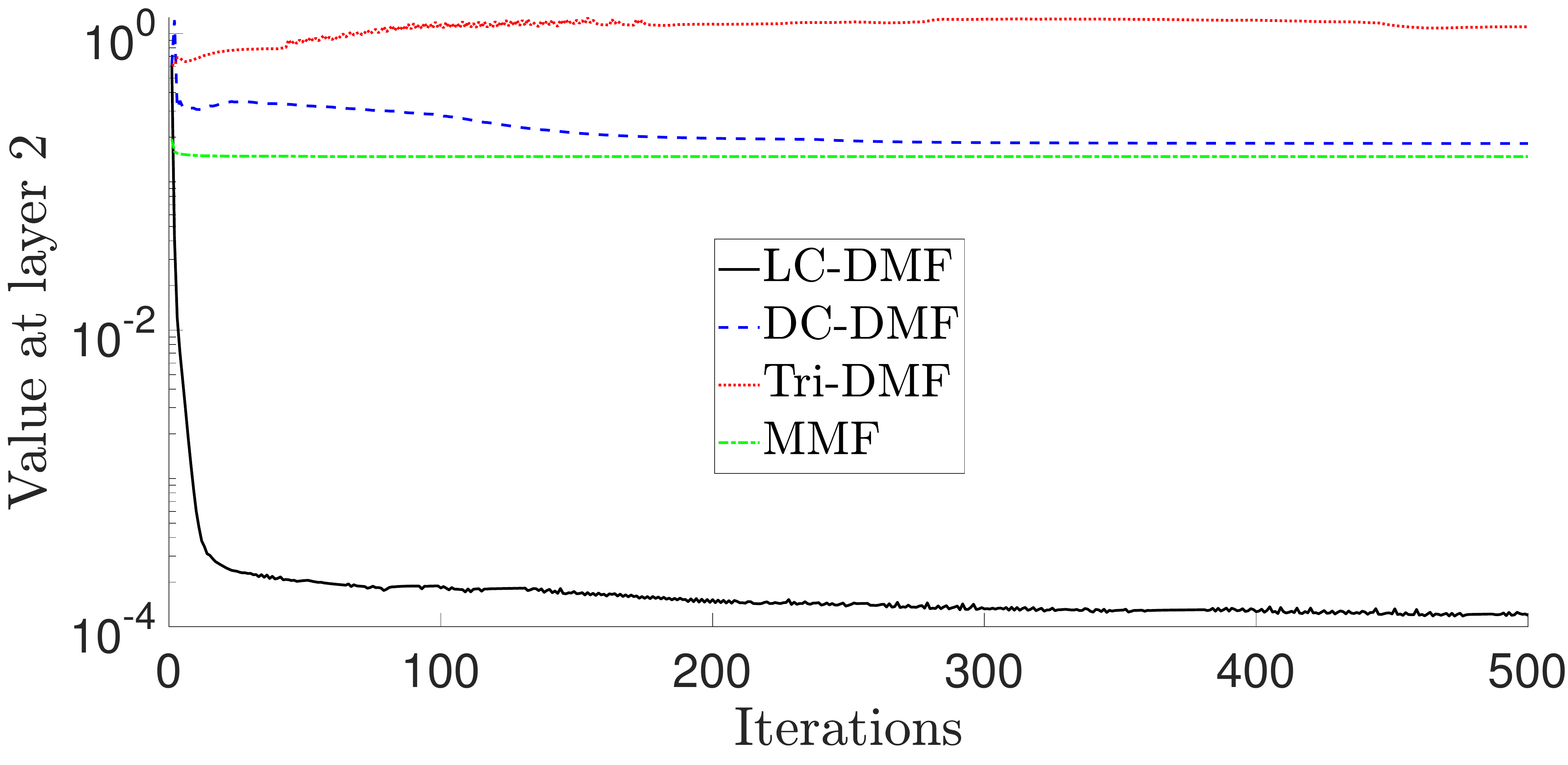}
  \label{fig:2B}}
  \hspace{1mm}
  \subfloat[][]{\hspace{-3mm}\includegraphics[width=0.7\textwidth]{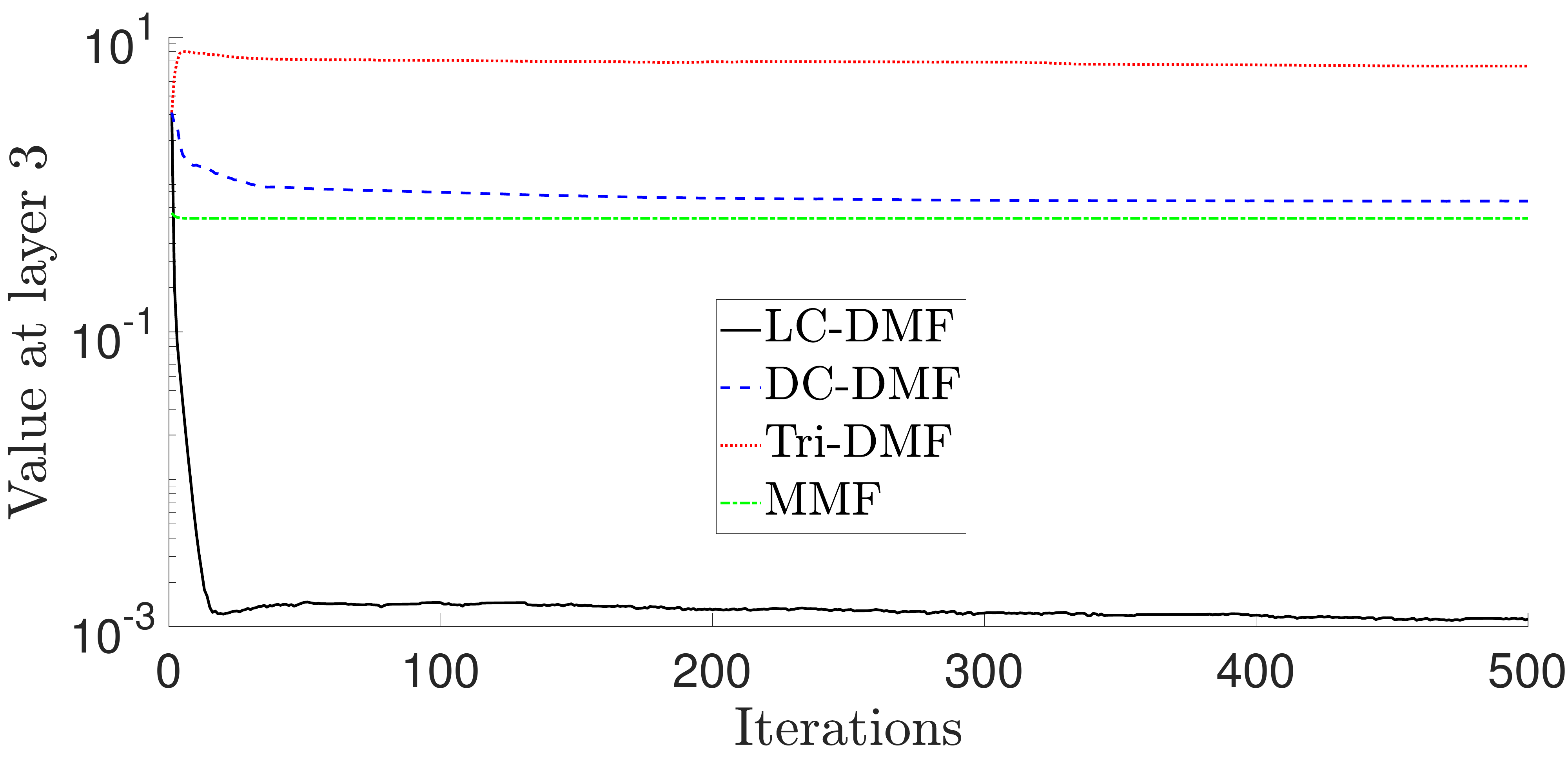}
  \label{fig:2C}}    
 
\caption{Layer-centric errors on the CBCL face data set, with $L=3$, $r_1=100$, $r_2=49$ and $r_3=25$ at the \protect\subref{fig:2A} first, 
\protect\subref{fig:2B} second, and \protect\subref{fig:2C} third layer.
}
\label{fig:err1}
\end{figure*}

\begin{figure}
  \centering
  \subfloat[][]{\includegraphics[width=0.7\textwidth]{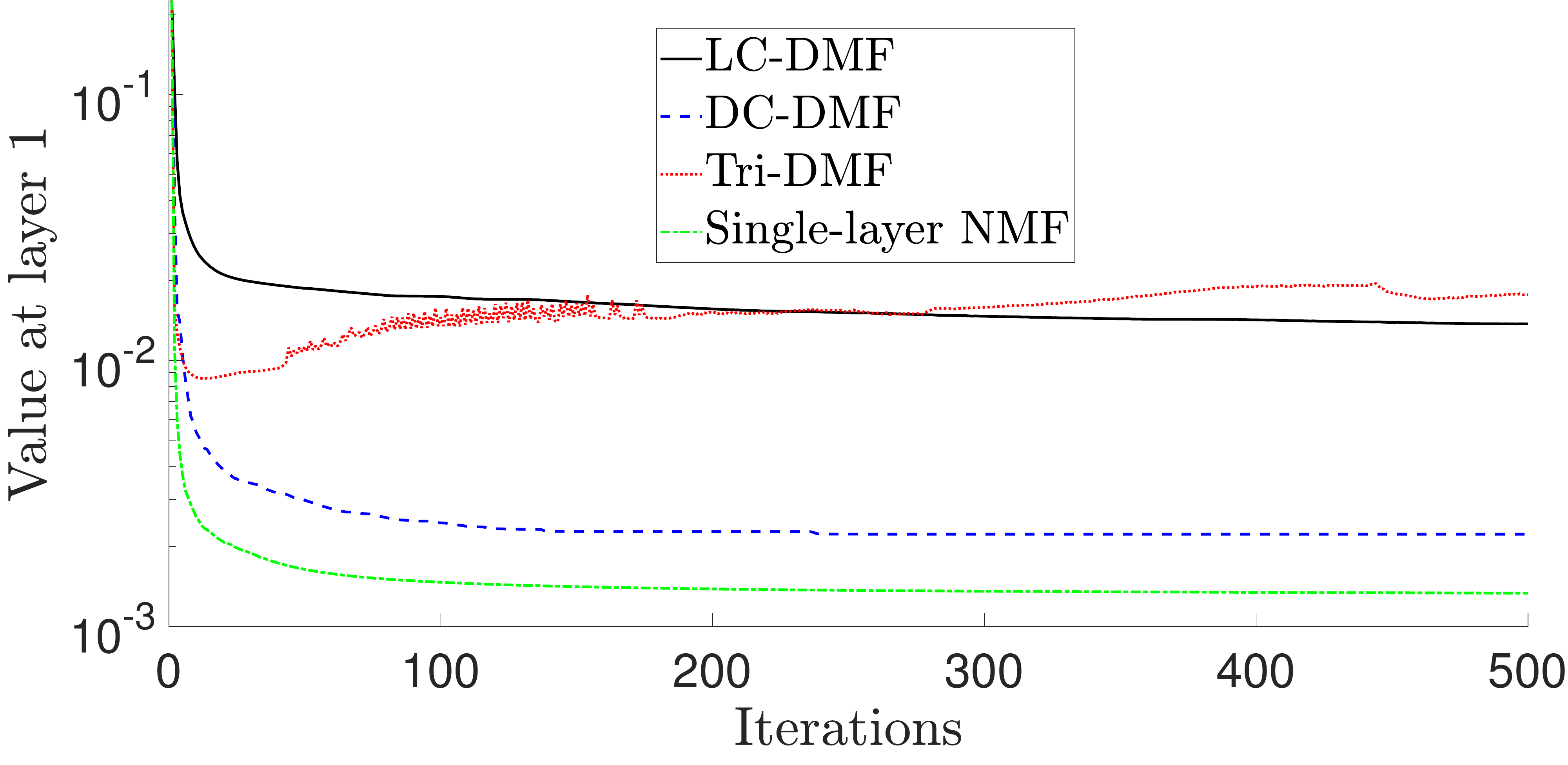}
  \label{fig:1A}}
  \hspace{3mm}
  \subfloat[][]{\includegraphics[width=0.7\textwidth]{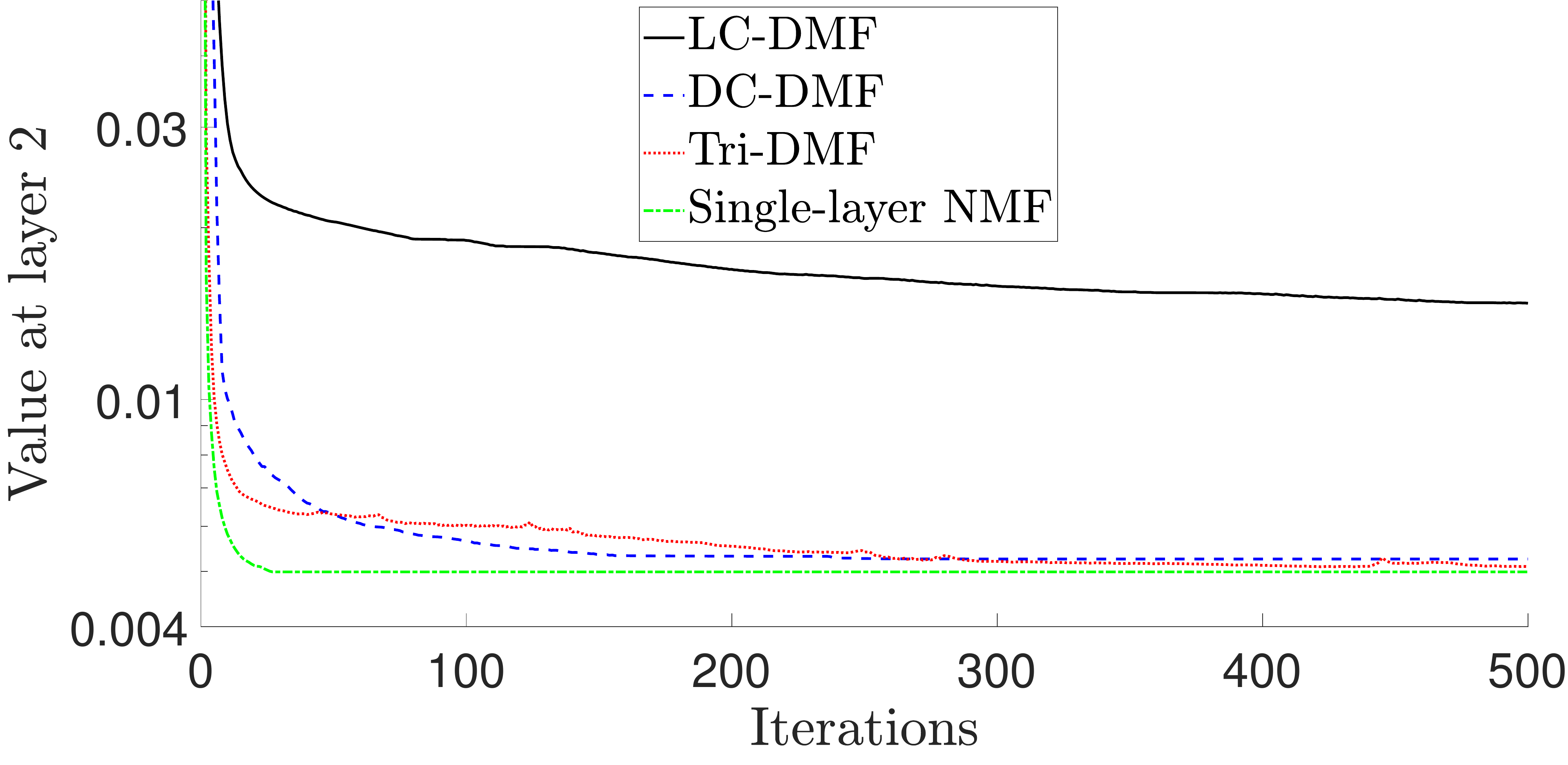}
  \label{fig:1B}}
  \hspace{3mm}
   \subfloat[][]{\includegraphics[width=0.7\textwidth]{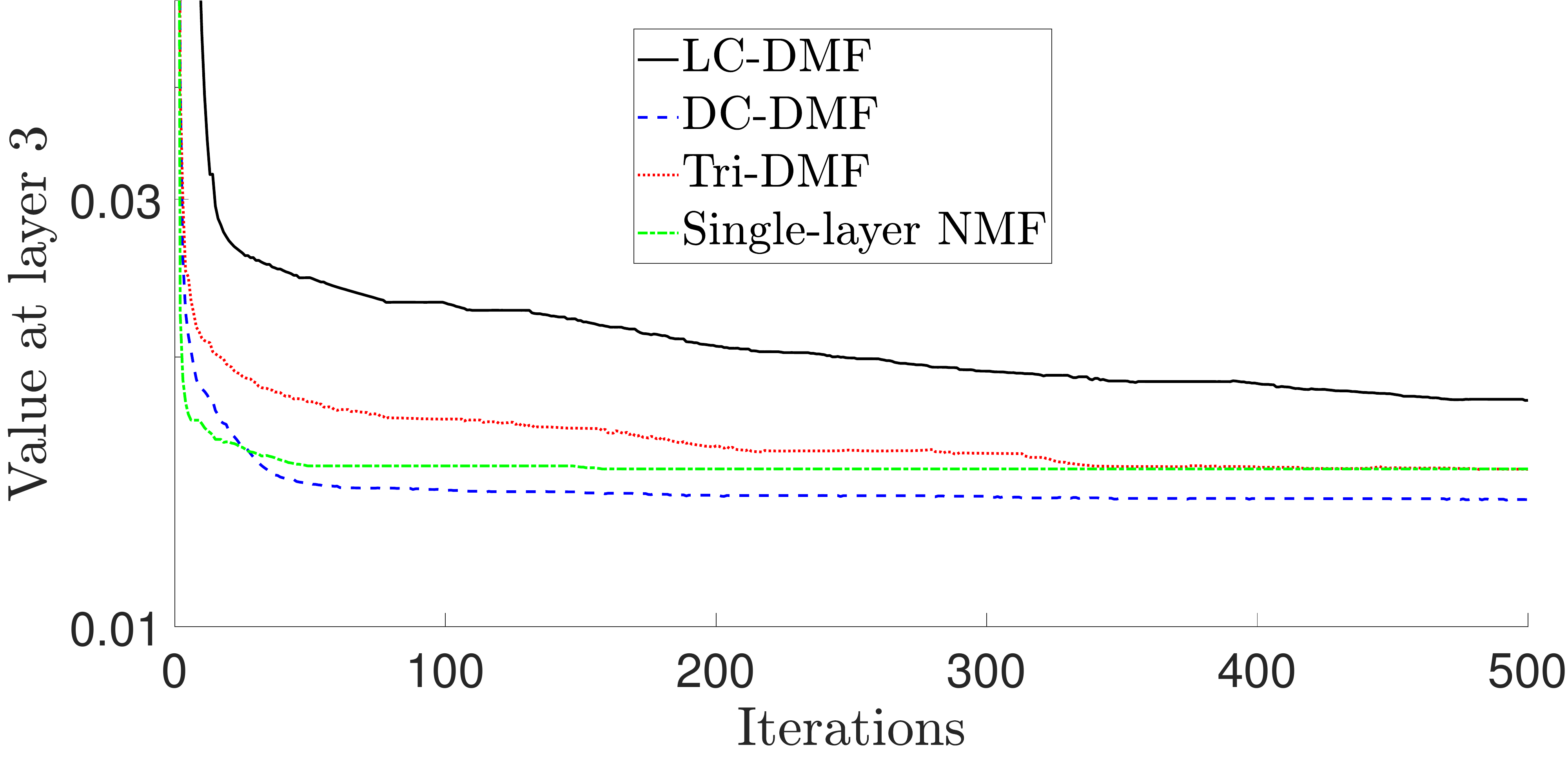}
  \label{fig:1C}}    
 
\caption{Data-centric errors on the CBCL face data set, with $L=3$, $r_1=100$, $r_2=49$ and $r_3=25$ at the \protect\subref{fig:1A} first, 
\protect\subref{fig:1B} second, and \protect\subref{fig:1C} third layer.
} 
\label{fig:err2}
\end{figure}

To compare quantitatively these methods, we plot the evolution of several loss functions along the $500$ iterations. More precisely, Fig.~\ref{fig:err1} shows the relative layer-centric errors $\frac{\|W_{l-1}-W_lH_l\|_F^2}{\|W_{l-1}^{(0)}\|_F^2}$ for $l=1, 2, 3$, with $W_0=X$, for all methods. Fig.~\ref{fig:err2} shows the relative data-centric errors $\frac{\|X-W_lH_l...H_1\|_F^2}{\|X\|_F^2}$ for $l=1, 2, 3$. Note that the first layer errors are both equal to $\frac{\|X-W_1H_1\|_F^2}{\|X\|_F^2}$ and appear in all global loss functions except \mbox{Tri-DMF}.  

This experiment confirms the advantage of \mbox{DC-DMF} and \mbox{LC-DMF} over MMF and \mbox{Tri-DMF}: 
\begin{itemize}
    \item At the second and third layers, \mbox{DC-DMF} produces the lowest \mbox{data-centric} errors while \mbox{LC-DMF} produces the lowest \mbox{layer-centric} errors. Note however that \mbox{DC-DMF} has slightly larger errors than single-layer NMF at the first two layers, while it has a lower error at the third layer. 
    This is expected since \mbox{DC-DMF} optimizes all layers simultaneously while single-layer NMF simply performs independent NMFs, that is,  \mbox{DC-DMF}  provides a hierarchy of intricated features.

    \item MMF has much higher relative errors than \mbox{LC-DMF} at the second and third layers (MMF is above~$10^{-1}$ while \mbox{LC-DMF} is below or about~$10^{-3}$). 
   This comes from the sequential optimization procedure of MMF. 
   More precisely, the factors of the first layer, $W_1$ and $H_1$, are first extracted, then those of the second layer, and so on, without any possibility of ``backpropagation'', see Section~\ref{sec:meth} for details.
   
   \item The first-layer error of \mbox{Tri-DMF} oscillates and does not converge (in fact, it appears to diverge), which is an expected consequence of the  different loss functions minimized at each layer; see the discussion of Section~\ref{sec:meth} for more details. 
\end{itemize}

\section{Conclusion}
\label{sec:ccl}

In this paper, we have discussed the choice of the loss functions in deep MF models. Our motivation came from the fact that the mainstream framework proposed by Trigeorgis et al.~\cite{trigeorgis2016deep}, namely the loss function~\eqref{eq:lossF} and the corresponding Algorithm~\ref{algo2}, is not consistent as it optimizes different losses at different layers. In fact, we have shown that this approach leads to poor feature extraction, and divergence of the loss functions. 
We have therefore proposed two loss functions for deep MF that naturally weighs the different layers, namely a layer-centric loss and a data-centric loss, 
see~\eqref{eq:loss1} and~\eqref{eq:loss2}, along with a simple block coordinate descent framework to compute the factors in the corresponding deep MF models. 
We focused on nonnegativity and sparsity constraints, and proposed a new deep MF model relying on minimum-volume regularization. 
We showed through experiments on synthetic and real data that our weighted loss functions allow us to outperform sequential MF, single-layer MF and the mainstream deep MF model from~\cite{trigeorgis2016deep}, while offering flexibility when various constraints or regularizations are used.  
\revise{In particular, we recommend to use the loss function defined in~\eqref{eq:loss1}, that is, LC-DMF, 
to tackle deep MF as it performed best on average in our numerical experiments.   
Besides, it is computationally much cheaper than DC-DMF as the dimension reduces as the factorization unfolds, while DC-NMF uses the full data set at each layer (see Section~\ref{sec:lossfunc} for the details). 
However, the best alternative will depend on the application and the data set at hand, and it is hard to know in advance which model will be the most appropriate. This also depends on the goal of the end user: for example, if the goal is to minimize the error w.r.t.\ the input data at all layers, then DC-DMF should be preferred.} 

An important direction of research is to find clever ways of choosing and tuning the regularization parameters in our proposed loss functions.  Moreover, other loss functions could be investigated, as variants of the two proposed ones. An other perspective is to embed deep MF in a more powerful optimization framework such as TITAN~\cite{hien2020inertial} which has proven to be particularly efficient to tackle non-smooth non-convex problems, such as the grouped sparse variant of deep MF (see Section~\ref{subsec:sparseDNMF}). 
Studying the identifiability, that is, the uniqueness of the factors retrieved by these models is also an important issue, which has not been investigated much for deep MF, except for some quite specific settings \cite{malgouyres2016identifiability, zheng2021hierarchical}. Finally, applying the different models on other applications, such as topic modeling, would also be insightful.

\section*{Acknowledgement}

\revise{We thank the reviewers for their insightful comments that helped us improve the paper.} 

This work was supported by the Fonds de la Recherche Scientifique - FNRS (F.R.S.-FNRS) and the Fonds Wetenschappelijk Onderzoek - Vlaanderen (FWO) under EOS Project no O005318F-RG47, and by the Francqui foundation. 
Pierre De Handschutter is a research fellow of the F.R.S.-FNRS. The authors claim no conflict of interest. 

\appendix

\section{Abundance maps of the Urban hyperspectral image} \label{subsec:appendA}

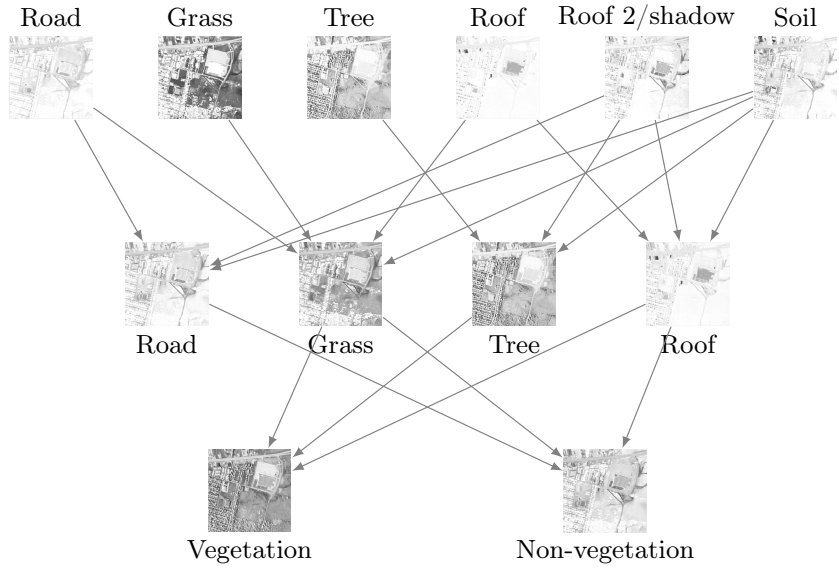
\begin{figure*}[h]
\centering
\scalebox{1.1}{
    \begin{tikzpicture}[thick]
  \begin{scope}[shift={(3.0,-0.5)},myBlock]
    \node [figNode={\includegraphics[width=1cm]{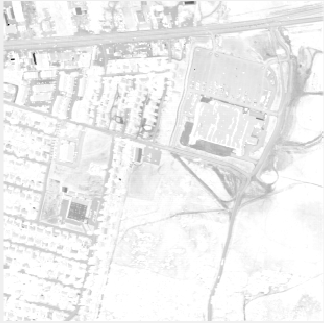}}] (A) at (-6.8,6) [label={[label distance=-0.3cm]above:\footnotesize Road}]{};
    \node[figNode={\includegraphics[width=1cm]{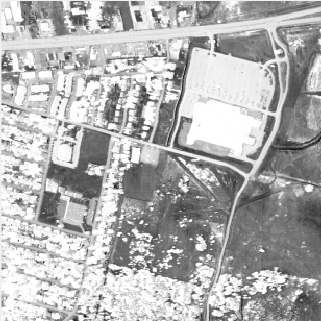}}](B) at (-5,6) [label={[label distance=-0.3cm]above:\footnotesize Grass}]{};
    \node[figNode={\includegraphics[width=1cm]{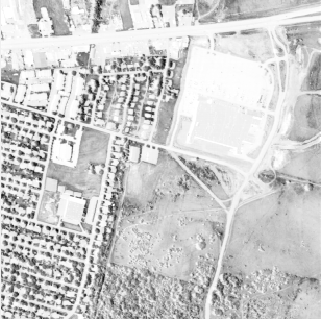}}] (C) at (-3.2,6)[label={[label distance=-0.3cm]above:\footnotesize Tree}] {};
    \node[figNode={\includegraphics[width=1cm]{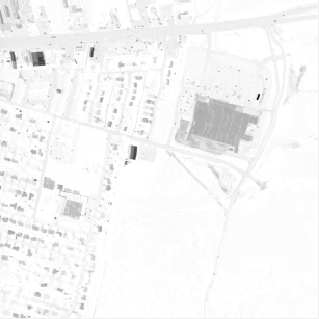}}] (D) at (-1.4,6) [label={[label distance=-0.3cm]above:\footnotesize Roof}]{};
    \node[figNode={\includegraphics[width=1cm]{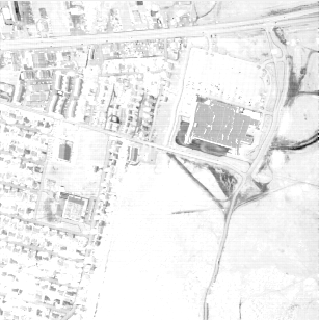}}] (E) at (0.4,6) [label={[label distance=-0.3cm]above:\footnotesize Roof 2/shadow}]{};
    \node[figNode={\includegraphics[width=1cm]{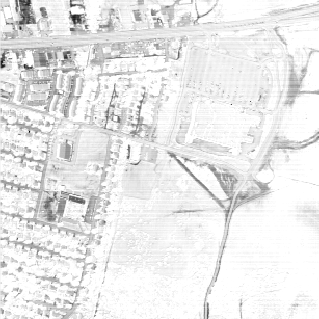}}] (F) at (2.2,6) [label={[label distance=-0.3cm]above:\footnotesize Soil}]{};

    \node[figNode={\includegraphics[width=1cm]{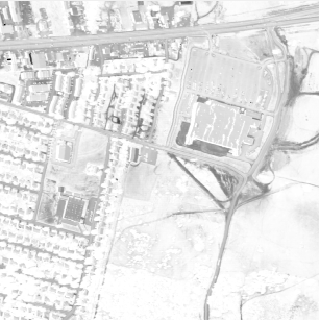}}] (H) at (-5.4,3.5) [label={[label distance=-0.3cm]below:\footnotesize Road}]{};
    \node[figNode={\includegraphics[width=1cm]{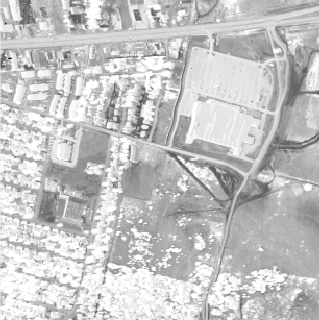}}] (I) at (-3.3,3.5) [label={[label distance=-0.3cm]below:\footnotesize Grass}]{};
    \node[figNode={\includegraphics[width=1cm]{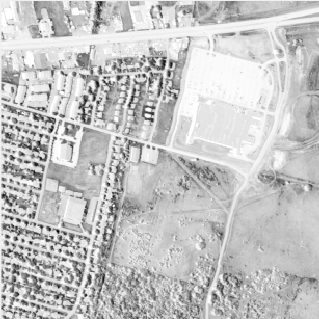}}] (J) at (-1.2,3.5) [label={[label distance=-0.3cm]below:\footnotesize Tree}]{};
    \node[figNode={\includegraphics[width=1cm]{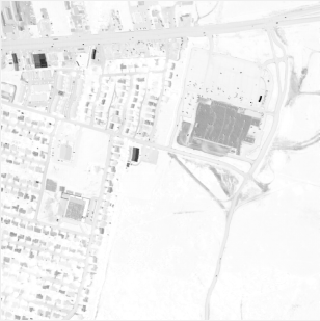}}] (K) at (0.9,3.5) [label={[label distance=-0.3cm]below:\footnotesize Roof}]{};
    
    \node[figNode={\includegraphics[width=1cm]{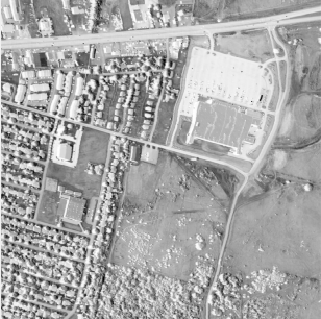}}] (L) at (-4.4,1) [label={[label distance=-0.3cm]below:\footnotesize Vegetation}]{};
    \node[figNode={\includegraphics[width=1cm]{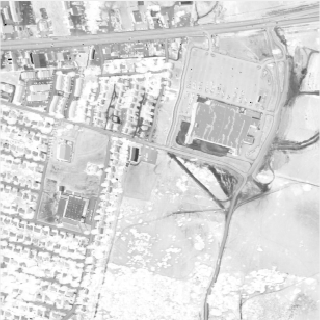}}] (M) at (-0.1,1 ) [label={[label distance=-0.3cm]below:\footnotesize Non-vegetation}]{};
  \end{scope}
  
  \begin{scope}[mySimpleArrow]
    \path[->, thin] (A) -- (H);
    \path[->, thin] (A) -- (I);
    \path[->, thin] (B) -- (I);
    \path[->, thin] (C) -- (J);
    \path[->, thin] (D) -- (I);
    \path[->, thin] (D) -- (K);
    \path[->, thin] (E) -- (H);
    \path[->, thin] (E) -- (J);
    \path[->, thin] (E) -- (K);
    \path[->, thin] (F) -- (H);
   \path[->, thin] (F) -- (I);
   \path[->, thin] (F) -- (J);
   \path[->, thin] (F) -- (K);
   \path[->, thin] (H) -- (M);
   \path[->, thin] (I) -- (L);
   \path[->, thin] (I) -- (M);
   \path[->, thin] (J) -- (L);
   \path[->, thin] (K) -- (L);
   \path[->, thin] (K) -- (M);
  \end{scope}
\end{tikzpicture}}
\vspace{-3mm}
    \caption{Hierarchy of features extracted by MMF on the Urban hyperspectral image with $L=3$ layers, $r_1=6$, $r_2=4$, $r_3=2$. \label{fig:Abund_MMF}}
\end{figure*}

\begin{figure*}
\centering
\scalebox{1.1}{
    \begin{tikzpicture}[thick]
  \begin{scope}[shift={(3.0,-0.5)},myBlock]
   \node [figNode={\includegraphics[width=1cm]{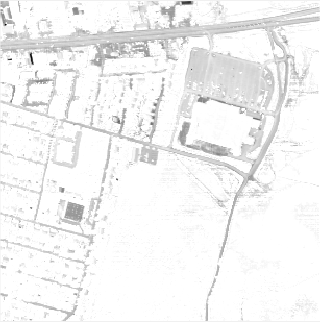}}] (A) at (-6.8,6) [label={[label distance=-0.3cm]above:\footnotesize Road}]{};
    \node[figNode={\includegraphics[width=1cm]{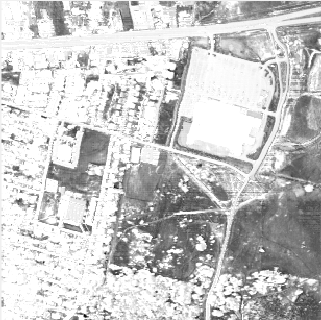}}](B) at (-5,6) [label={[label distance=-0.3cm]above:\footnotesize Grass}]{};
    \node[figNode={\includegraphics[width=1cm]{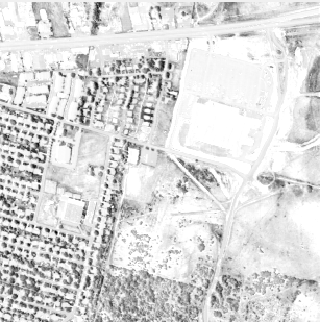}}] (C) at (-3.2,6)[label={[label distance=-0.3cm]above:\footnotesize Tree}] {};
    \node[figNode={\includegraphics[width=1cm]{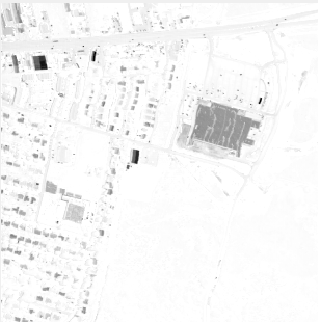}}] (D) at (-1.4,6) [label={[label distance=-0.3cm]above:\footnotesize Roof}]{};
    \node[figNode={\includegraphics[width=1cm]{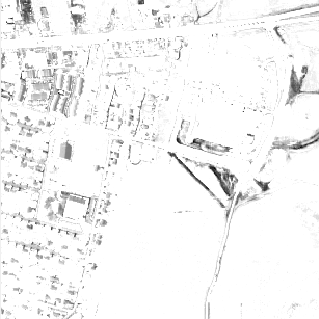}}] (E) at (0.4,6) [label={[label distance=-0.3cm]above:\footnotesize Roof 2/shadow}]{};
    \node[figNode={\includegraphics[width=1cm]{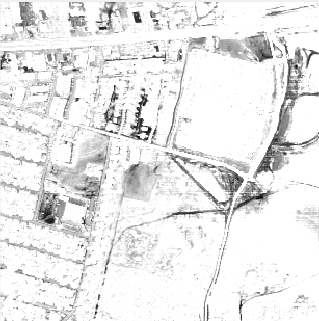}}] (F) at (2.2,6) [label={[label distance=-0.3cm]above:\footnotesize Soil}]{};

    \node[figNode={\includegraphics[width=1cm]{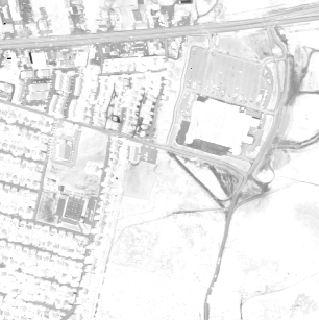}}] (H) at (-5.4,3.5) [label={[label distance=-0.3cm]below:\footnotesize Road}]{};
    \node[figNode={\includegraphics[width=1cm]{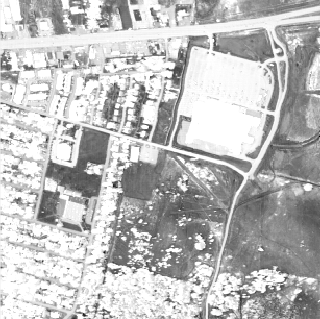}}] (I) at (-3.3,3.5) [label={[label distance=-0.3cm]below:\footnotesize Grass}]{};
    \node[figNode={\includegraphics[width=1cm]{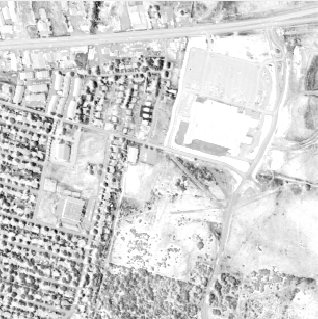}}] (J) at (-1.2,3.5) [label={[label distance=-0.3cm]below:\footnotesize Tree}]{};
    \node[figNode={\includegraphics[width=1cm]{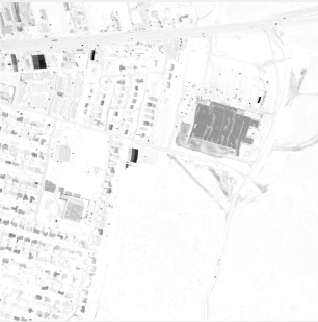}}] (K) at (0.9,3.5) [label={[label distance=-0.3cm]below:\footnotesize Roof}]{};
    
    \node[figNode={\includegraphics[width=1cm]{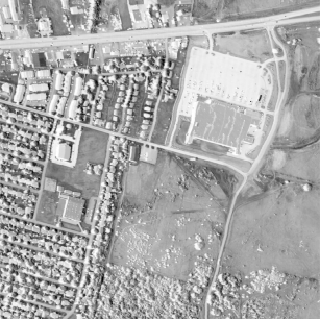}}] (L) at (-4.4,1) [label={[label distance=-0.3cm]below:\footnotesize Vegetation}]{};
    \node[figNode={\includegraphics[width=1cm]{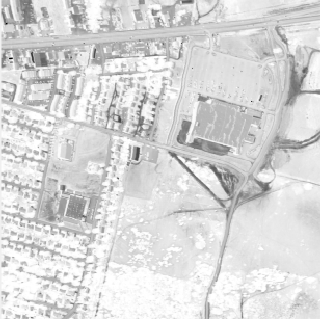}}] (M) at (-0.1,1 ) [label={[label distance=-0.3cm]below:\footnotesize Non-vegetation}]{};
  \end{scope}
  
  \begin{scope}[mySimpleArrow]
    \path[->, thin] (A) -- (H);
    \path[->, thin] (A) -- (J);
    \path[->, thin] (B) -- (I);
    \path[->, thin] (C) -- (J);
    \path[->, thin] (D) -- (K);
    \path[->, thin] (E) -- (H);
    \path[->, thin] (E) -- (J);
    \path[->, thin] (E) -- (K);
    \path[->, thin] (F) -- (H);
    \path[->, thin] (F) -- (I);
    \path[->, thin] (H) -- (M);
   \path[->, thin] (I) -- (L);
   \path[->, thin] (I) -- (M);
   \path[->, thin] (J) -- (L);
   \path[->, thin] (K) -- (L);
   \path[->, thin] (K) -- (M);
  \end{scope}
\end{tikzpicture}}
\vspace{-3mm}
    \caption{Hierarchy of features extracted by LC-DMF on the Urban hyperspectral image with $L=3$ layers, $r_1=6$, $r_2=4$, $r_3=2$.}
    \label{fig:Abund_LC}
\end{figure*}

\begin{figure*}
\centering
\scalebox{1.1}{
    \begin{tikzpicture}[thick]
  \begin{scope}[shift={(3.0,-0.5)},myBlock]
    \node [figNode={\includegraphics[width=1cm]{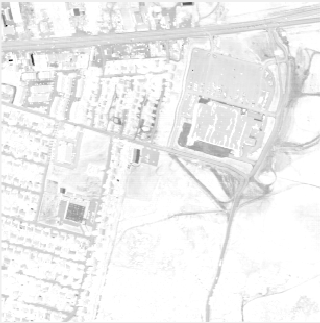}}] (A) at (-6.8,6) [label={[label distance=-0.3cm]above:\footnotesize Road}]{};
    \node[figNode={\includegraphics[width=1cm]{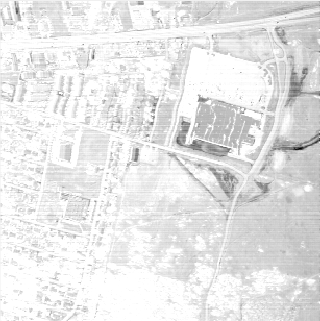}}](B) at (-5,6) [label={[label distance=-0.3cm]above:\footnotesize Grass}]{};
    \node[figNode={\includegraphics[width=1cm]{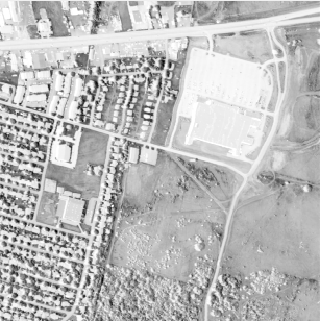}}] (C) at (-3.2,6)[label={[label distance=-0.3cm]above:\footnotesize Tree}] {};
    \node[figNode={\includegraphics[width=1cm]{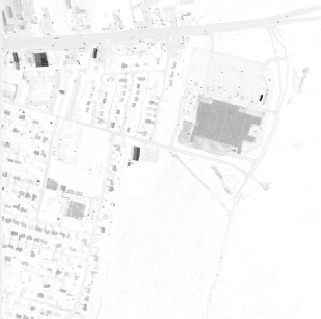}}] (D) at (-1.4,6) [label={[label distance=-0.3cm]above:\footnotesize Roof}]{};
    \node[figNode={\includegraphics[width=1cm]{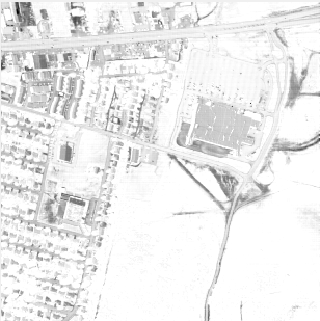}}] (E) at (0.4,6) [label={[label distance=-0.3cm]above:\footnotesize Roof 2/shadow}]{};
    \node[figNode={\includegraphics[width=1cm]{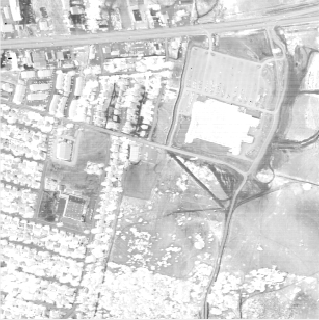}}] (F) at (2.2,6) [label={[label distance=-0.3cm]above:\footnotesize Soil}]{};

    \node[figNode={\includegraphics[width=1cm]{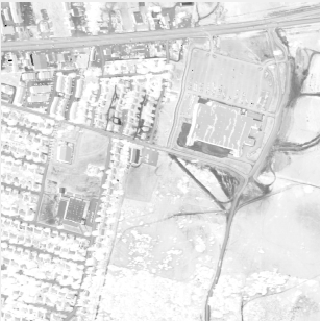}}] (H) at (-5.4,3.5) [label={[label distance=-0.3cm]below:\footnotesize Road}]{};
    \node[figNode={\includegraphics[width=1cm]{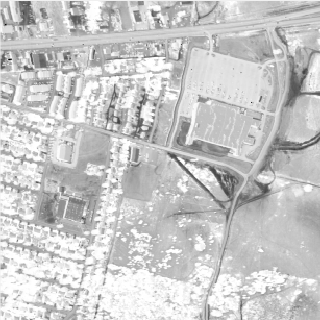}}] (I) at (-3.3,3.5) [label={[label distance=-0.3cm]below:\footnotesize Grass}]{};
    \node[figNode={\includegraphics[width=1cm]{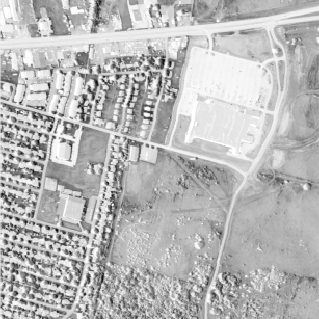}}] (J) at (-1.2,3.5) [label={[label distance=-0.3cm]below:\footnotesize Tree}]{};
    \node[figNode={\includegraphics[width=1cm]{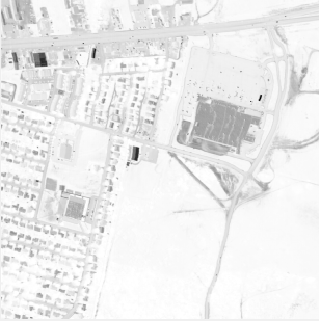}}] (K) at (0.9,3.5) [label={[label distance=-0.3cm]below:\footnotesize Roof}]{};
    
    \node[figNode={\includegraphics[width=1cm]{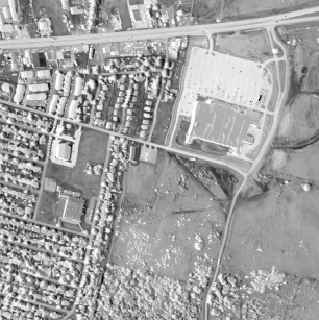}}] (L) at (-4.4,1) [label={[label distance=-0.3cm]below:\footnotesize Vegetation}]{};
    \node[figNode={\includegraphics[width=1cm]{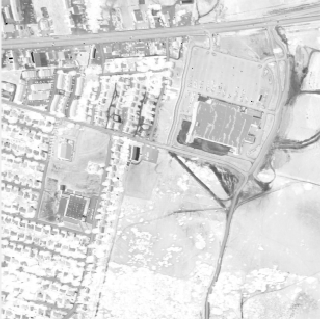}}] (M) at (-0.1,1) [label={[label distance=-0.3cm]below:\footnotesize Non-vegetation}]{};
  \end{scope}
  
  \begin{scope}[mySimpleArrow]
    \path[->, thin] (A) -- (H);
    \path[->, thin] (A) -- (K);
    \path[->, thin] (B) -- (I);
    \path[->, thin] (B) -- (H);
    \path[->, thin] (B) -- (K);
    \path[->, thin] (C) -- (J);
    \path[->, thin] (D) -- (I);
    \path[->, thin] (D) -- (K);
     \path[->, thin] (E) -- (H);
    \path[->, thin] (E) -- (J);
    \path[->, thin] (E) -- (K);
    \path[->, thin] (F) -- (H);
    \path[->, thin] (F) -- (I);
    \path[->, thin] (H) -- (M);
   \path[->, thin] (I) -- (L);
   \path[->, thin] (I) -- (M);
   \path[->, thin] (J) -- (L);
   \path[->, thin] (K) -- (L);
   \path[->, thin] (K) -- (M);
  \end{scope}
\end{tikzpicture}}
\vspace{-3mm}
    \caption{Hierarchy of features extracted by DC-DMF on the Urban hyperspectral image with $L=3$ layers, $r_1=6$, $r_2=4$, $r_3=2$.}
    \label{fig:Abund_DC}
\end{figure*}

\begin{figure*}
\centering
\scalebox{1.1}{
    \begin{tikzpicture}[thick]
  \begin{scope}[shift={(3.0,-0.5)},myBlock]
    \node [figNode={\includegraphics[width=1cm]{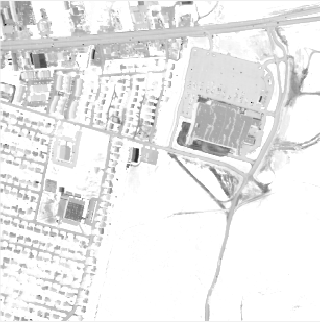}}] (A) at (-6.8,6) [label={[label distance=-0.3cm]above:\footnotesize Road}]{};
    \node[figNode={\includegraphics[width=1cm]{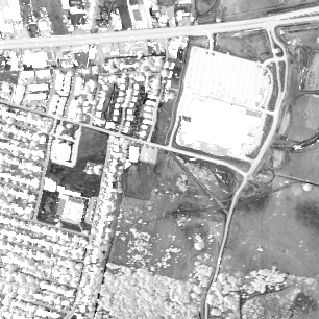}}](B) at (-5,6) [label={[label distance=-0.3cm]above:\footnotesize Grass}]{};
    \node[figNode={\includegraphics[width=1cm]{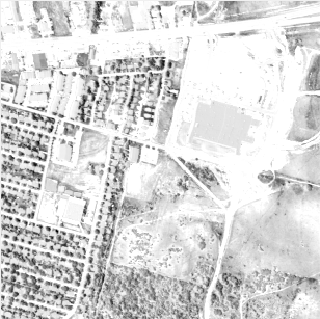}}] (C) at (-3.2,6)[label={[label distance=-0.3cm]above:\footnotesize Tree}] {};
    \node[figNode={\includegraphics[width=1cm]{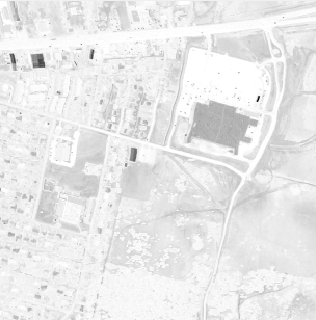}}] (D) at (-1.4,6) [label={[label distance=-0.3cm]above:\footnotesize Roof}]{};
    \node[figNode={\includegraphics[width=1cm]{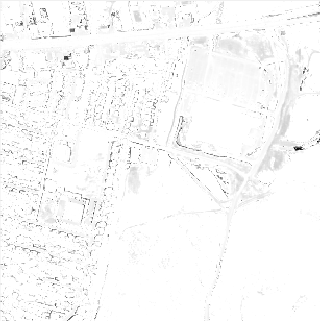}}] (E) at (0.4,6) [label={[label distance=-0.3cm]above:\footnotesize Roof 2/shadow}]{};
    \node[figNode={\includegraphics[width=1cm]{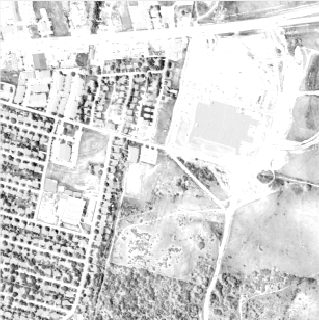}}] (F) at (2.2,6) [label={[label distance=-0.3cm]above:\footnotesize Soil}]{};

    \node[figNode={\includegraphics[width=1cm]{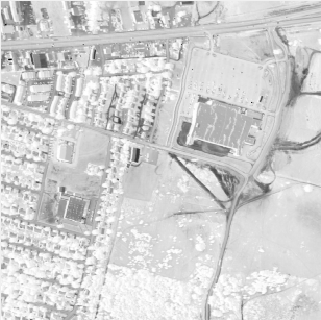}}] (H) at (-5.4,3.5) [label={[label distance=-0.3cm]below:\footnotesize Road}]{};
    \node[figNode={\includegraphics[width=1cm]{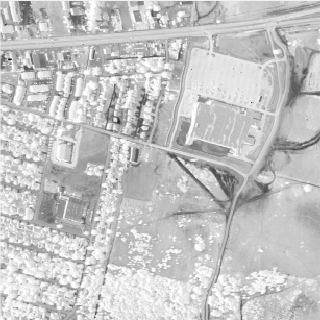}}] (I) at (-3.3,3.5) [label={[label distance=-0.3cm]below:\footnotesize Grass}]{};
    \node[figNode={\includegraphics[width=1cm]{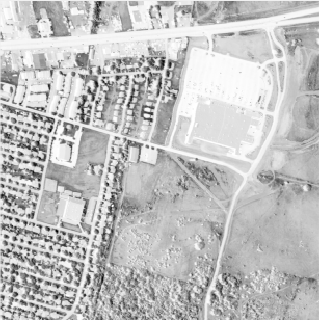}}] (J) at (-1.2,3.5) [label={[label distance=-0.3cm]below:\footnotesize Tree}]{};
    \node[figNode={\includegraphics[width=1cm]{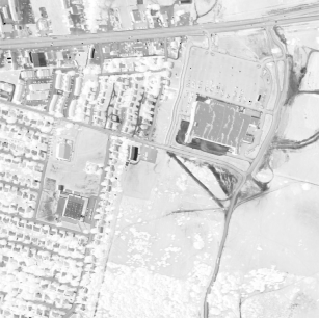}}] (K) at (0.9,3.5) [label={[label distance=-0.3cm]below:\footnotesize Roof}]{};
    
    \node[figNode={\includegraphics[width=1cm]{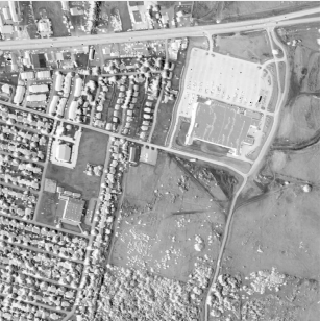}}] (L) at (-4.4,1) [label={[label distance=-0.3cm]below:\footnotesize Vegetation}]{};
    \node[figNode={\includegraphics[width=1cm]{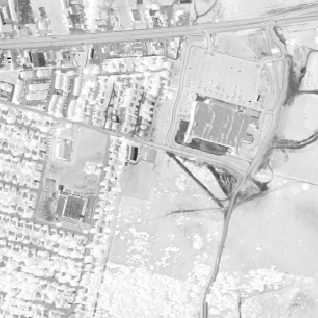}}] (M) at (-0.1,1 ) [label={[label distance=-0.3cm]below:\footnotesize Non-vegetation}]{};
  \end{scope}
  
  \begin{scope}[mySimpleArrow]
    \path[->, thin] (A) -- (H);
    \path[->, thin] (A) -- (I);
    \path[->, thin] (A) -- (K);
     \path[->, thin] (B) -- (H);
    \path[->, thin] (B) -- (I);
    \path[->, thin] (B) -- (J);
    \path[->, thin] (C) -- (H);
    \path[->, thin] (C) -- (I);
    \path[->, thin] (C) -- (J);
    \path[->, thin] (D) -- (H);
    \path[->, thin] (D) -- (I);
    \path[->, thin] (D) -- (J);
    \path[->, thin] (D) -- (K);
     \path[->, thin] (E) -- (H);
    \path[->, thin] (E) -- (I);
    \path[->, thin] (E) -- (J);
    \path[->, thin] (E) -- (K);
     \path[->, thin] (F) -- (H);
    \path[->, thin] (F) -- (I);
    \path[->, thin] (F) -- (J);
    \path[->, thin] (F) -- (K);
    
    \path[->, thin] (H) -- (M);
     \path[->, thin] (H) -- (L);
   \path[->, thin] (I) -- (L);
   \path[->, thin] (I) -- (M);
   \path[->, thin] (J) -- (L);
   \path[->, thin] (K) -- (M);
  \end{scope}
\end{tikzpicture}}
\vspace{-3mm}
    \caption{Hierarchy of features extracted by Tri-DMF on the Urban hyperspectral image with $L=3$ layers, $r_1=6$, $r_2=4$, $r_3=2$.}
    \label{fig:Abund_Tri}
\end{figure*}

\begin{figure*} [t]
    \centering
  \subfloat[Road]{\includegraphics[scale=0.3]{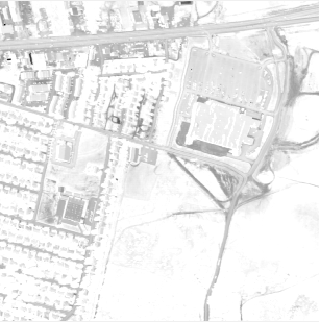}
  \label{fig:single_road}}
  \hspace{6mm}
   \subfloat[Grass]{\includegraphics[scale=0.3]{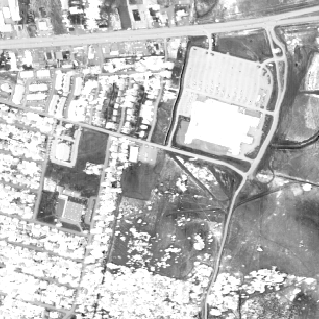}
  \label{fig:single_grass}}
  \hspace{6mm}
   \subfloat[Tree]{\includegraphics[scale=0.3]{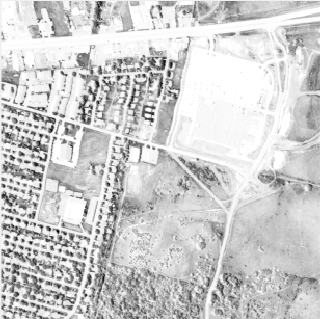}
  \label{fig:single_tree}} 
  \hspace{6mm}
   \subfloat[Roof]{\includegraphics[scale=0.3]{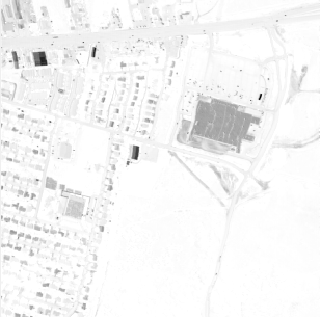}
  \label{fig:single_roof}} 
    \caption{Features extracted by single-layer NMF with $r=4$ on the Urban hyperspectral image.}
    \label{fig:single4}
\end{figure*}

\begin{figure*} [!h]
    \centering
  \subfloat[Vegetation]{\includegraphics[scale=0.45]{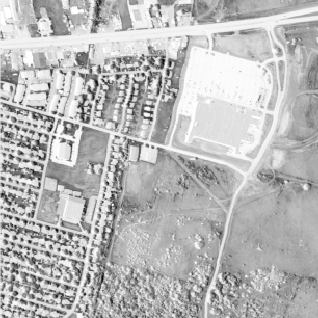}
  \label{fig:single_veg}}
  \hspace{10mm}
   \subfloat[Non-vegetation]{\includegraphics[scale=0.45]{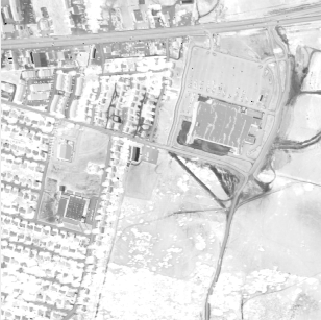}
  \label{fig:single_hum}}
      \caption{Features extracted by single-layer NMF with $r=2$ on the Urban hyperspectral image. \label{fig:single2}} 
\end{figure*}

Figures~\ref{fig:Abund_MMF}-\ref{fig:single2} provide the abundance maps obtained with MMF, LC-DMF, DC-DMF, Tri-DMF and single-layer NMF applied on the Urban image for factorizations of depth $L=3$, with $r_1=6$, $r_2=4$, $r_3=2$ (see Section~\ref{subsubsec:hyper}).

\section{Facial features of the CBCL dataset} \label{subsec:appendB}

In this section, we present the facial features extracted by \mbox{LC-DMF}, \mbox{DC-DMF}, \mbox{Tri-DMF} and \mbox{single-layer} NMF with grouped sparsity constraints on the CBCL dataset (see Section~\ref{subsubsec:faces}). 
\begin{figure*}[!h]
  \centering
  \subfloat[][]{\includegraphics[width=0.29\textwidth]{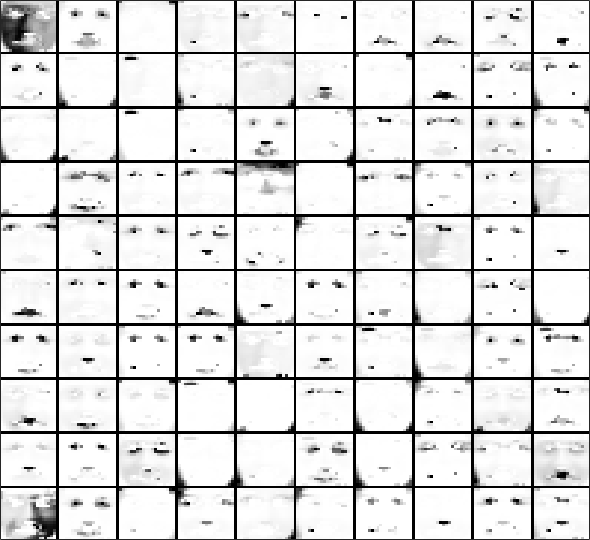}
  \label{fig:G1}}
  \hspace{4mm}
   \subfloat[][]{\includegraphics[width=0.29\textwidth]{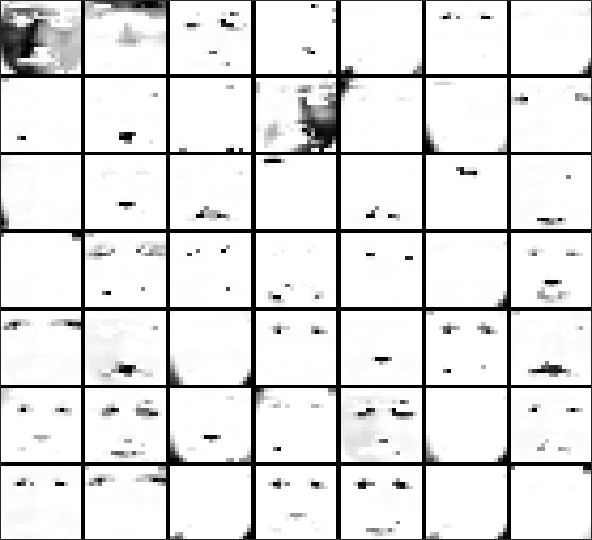}
  \label{fig:G2}}
  \hspace{4mm}
   \subfloat[][]{\includegraphics[width=0.29\textwidth]{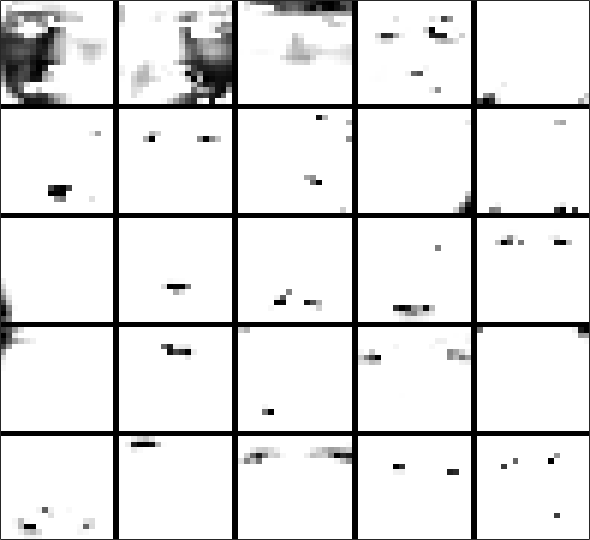}
  \label{fig:G3}}

\caption{Features extracted by LC-DMF on the CBCL face data set, with $L=3$, $r_1=100$, $r_2=49$, and $r_3=25$. Each image contains the features extracted at a layer: \protect\subref{fig:G1} first layer $W_1$, 
\protect\subref{fig:G2} second layer $W_2$, and \protect\subref{fig:G3} third layer $W_3$.
}
\label{fig:faces2}
\end{figure*}

\begin{figure*}[h]
  \centering
  \subfloat[][]{\includegraphics[width=0.29\textwidth]{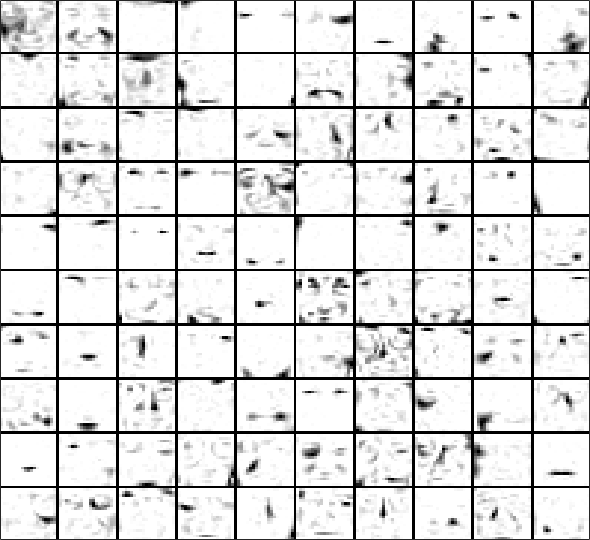}
  \label{fig:H1}}
  \hspace{4mm}
   \subfloat[][]{\includegraphics[width=0.29\textwidth]{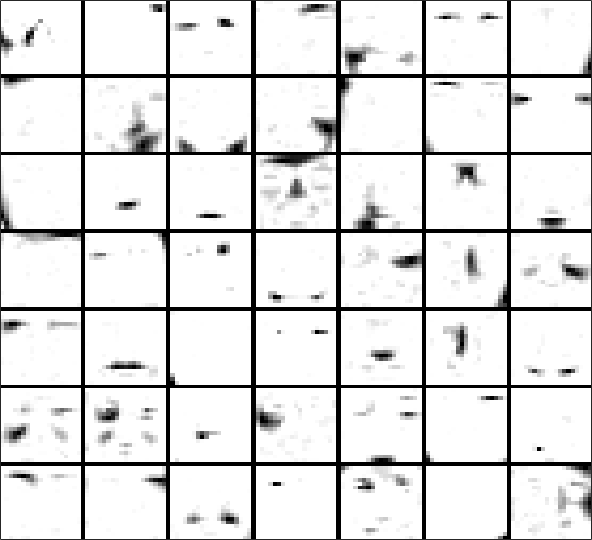}
  \label{fig:H2}}
  \hspace{4mm}
   \subfloat[][]{\includegraphics[width=0.29\textwidth]{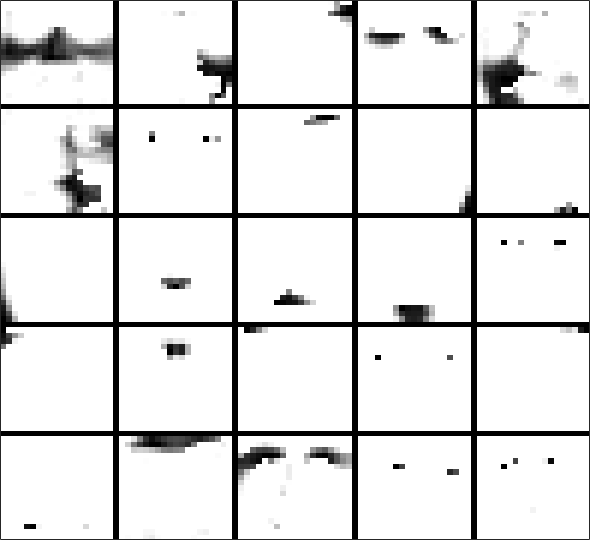}
  \label{fig:H3}}    

\caption{Features extracted by DC-DMF on the CBCL face data set, with $L=3$, $r_1=100$, $r_2=49$, and $r_3=25$. Each image contains the features extracted at a layer: \protect\subref{fig:H1} first layer $W_1$, 
\protect\subref{fig:H2} second layer $W_2$, and \protect\subref{fig:H3} third layer $W_3$.
}
\label{fig:faces3}
\end{figure*}

\begin{figure*}[h]
  \centering
  \subfloat[][]{\includegraphics[width=0.29\textwidth]{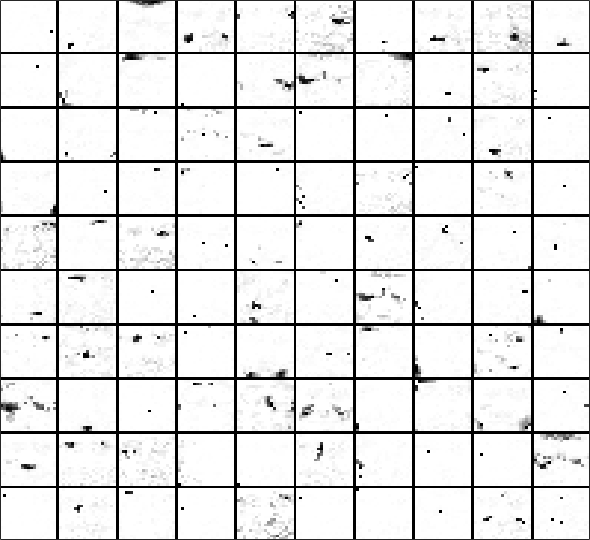}
  \label{fig:I1}}
  \hspace{4mm}
   \subfloat[][]{\includegraphics[width=0.29\textwidth]{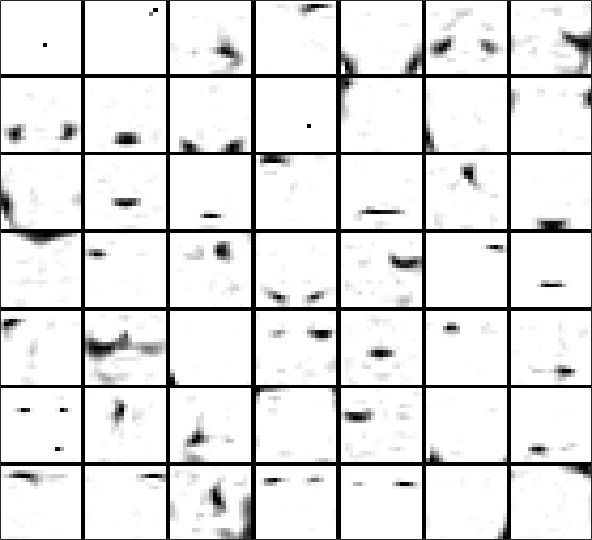}
  \label{fig:I2}}
  \hspace{4mm}
   \subfloat[][]{\includegraphics[width=0.29\textwidth]{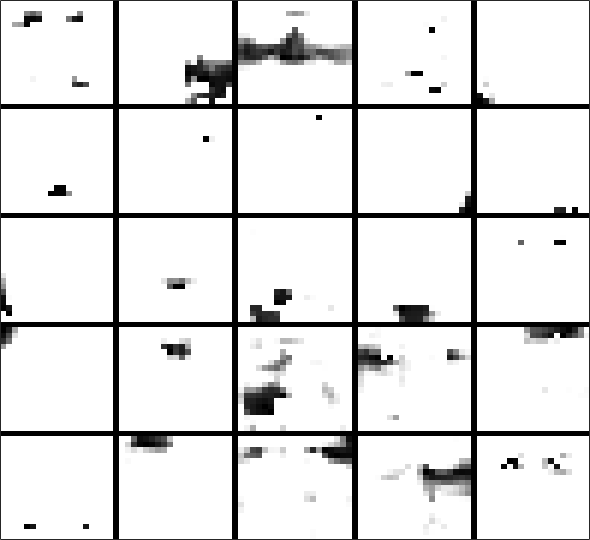}
  \label{fig:I3}}    

\caption{Features extracted by Tri-DMF on the CBCL face data set, with $L=3$, $r_1=100$, $r_2=49$, and $r_3=25$. Each image contains the features extracted at a layer: \protect\subref{fig:I1} first layer $W_1$, 
\protect\subref{fig:I2} second layer $W_2$, and \protect\subref{fig:I3} third layer $W_3$.
}
\label{fig:faces4}
\end{figure*}

\begin{figure*}
  \centering
  \subfloat[][]{\includegraphics[width=0.29\textwidth]{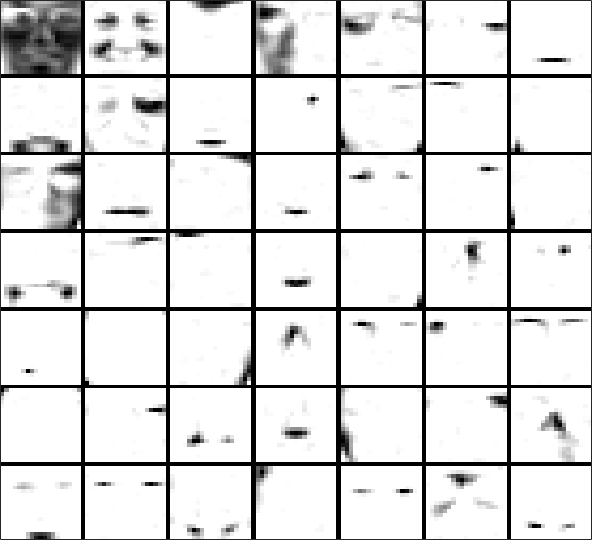}
  \label{fig:J1}}
  \hspace{4mm}
   \subfloat[][]{\includegraphics[width=0.29\textwidth]{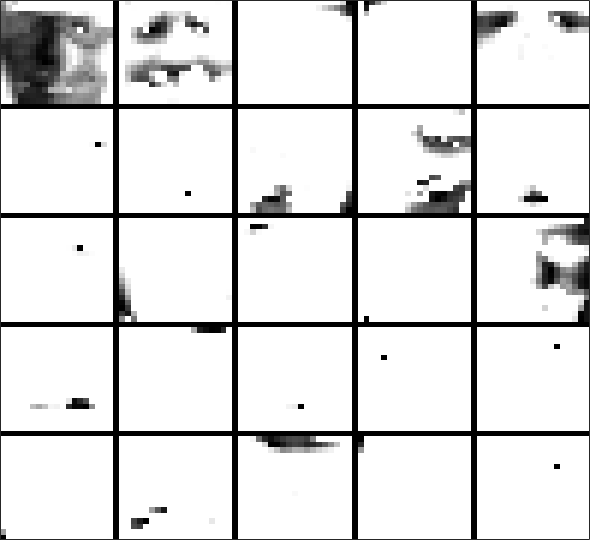}
  \label{fig:J2}}

\caption{Features extracted by single-layer NMF on the CBCL face data set, with \protect\subref{fig:J1}~$r=49$ 
\protect\subref{fig:J2} $r=25$.
}
\label{fig:faces5}
\end{figure*}


\small 
 \bibliographystyle{elsarticle-num}
 \clearpage 
 \bibliography{cas-refs}
 \pagebreak
% Biography Section
%\section*{ }

%\noindent \textbf{Pierre De Handschutter} received the Master's degree in computer science engineering from Université de Mons, Belgium, in 2019, where he is currently a F.R.S-FNRS research fellow with the Department of Mathematics and Operational Research. He started his Ph.D. in 2019 under the supervision of Prof. Nicolas Gillis. 
%\subsection*{  } % This subsection (with no heading) is added to give more space between two biographies
%\noindent \textbf{Nicolas Gillis} is Professor with the Department of Mathematics and Operational Research, University of Mons, Belgium. He is recipient of the Householder award and ERC starting grant. He is editor for IEEEE Transactions on Signal Processing, and SIAM Journals on Matrix Analysis and Applications and on Mathematics of Data Science. 
%

% \end{thebibliography}
\end{document}